\documentclass[10pt,letterpaper]{article}
\pdfoutput=1
\usepackage[top=0.85in,left=2.75in,footskip=0.75in]{geometry}
\usepackage{changepage}
\usepackage[utf8]{inputenc}
\usepackage{textcomp,marvosym}
\usepackage{fixltx2e}
\usepackage{amsmath,amssymb}
\usepackage{cite}
\usepackage{nameref,hyperref}
\usepackage[right]{lineno}
\usepackage{microtype}
\DisableLigatures[f]{encoding = *, family = * }
\usepackage{rotating}
\usepackage[aboveskip=1pt,labelfont=bf,labelsep=period,justification=raggedright,singlelinecheck=off]{caption}
\makeatletter
\renewcommand{\@biblabel}[1]{\quad#1.}
\makeatother
\date{}
\usepackage{lastpage,fancyhdr,graphicx}
\usepackage{epstopdf}
\fancyheadoffset[L]{2.25in}
\fancyfootoffset[L]{2.25in}

\usepackage{color}

\usepackage{rotating}

\usepackage{ifthen}
\newboolean{PrintVersion}
\setboolean{PrintVersion}{false} 

\usepackage{amsmath,amssymb,amstext} 
\let\origdoublepage\cleardoublepage
\newcommand{\clearemptydoublepage}{%
  \clearpage{\pagestyle{empty}\origdoublepage}}
\let\cleardoublepage\clearemptydoublepage

\usepackage{algorithm}
\usepackage{algpseudocode}
\usepackage{algorithmicx}
\usepackage{array}
\usepackage{multirow}
\usepackage[normalem]{ulem}
\usepackage{gensymb}
\usepackage{breakurl}

\begin{document}
\vspace*{0.35in}

\begin{flushleft}
{\Large
\textbf\newline{GeoTextTagger: High-Precision Location Tagging of Textual Documents using a Natural Language Processing Approach}
}
\newline
\\
Shawn Brunsting\textsuperscript{1},
Hans De Sterck*\textsuperscript{2},
Remco Dolman\textsuperscript{3},
Teun van Sprundel\textsuperscript{3}
\\
\bigskip
\bf{1} Centre for Computational Mathematics, University of Waterloo, Canada
\\
\bf{2} School of Mathematical Sciences, Monash University, Australia
\\
\bf{3} Spotzi Canada Ltd, Kitchener, Canada
\\
\bigskip

*hans.desterck@monash.edu

\end{flushleft}
\section*{Abstract}
Location tagging, also known as geotagging or geolocation, is the process of assigning geographical coordinates to input data. In this paper we present an algorithm for location tagging of textual documents.
Our approach makes use of previous work in natural language processing by using a state-of-the-art part-of-speech tagger and named entity recognizer to find blocks of text which may refer to locations. A knowledge base (OpenStreatMap) is then used to find a list of possible locations for each of these blocks of text. Finally, one location is chosen for each block of text by assigning distance-based scores to each location and repeatedly selecting the location and block of text with the best score.
We tested our geolocation algorithm with Wikipedia articles about topics with a well-defined geographical location that are geotagged by the articles' authors, where classification approaches have achieved median errors as low as 11 km.
However, the maximum accuracy of these approaches is limited by the class size, so future work may not yield significant improvement.
Our algorithm tags a location to each block of text that was identified as a possible location reference, meaning a text is typically assigned multiple tags. When we considered only the tag with the highest distance-based score, we achieved a 10th percentile error of 490 metres and median error of 54 kilometres on the Wikipedia dataset we used. When we considered the five location tags with the greatest scores, we found that 50\% of articles were assigned at least one tag within 8.5 kilometres of the article's author-assigned true location. We also tested our approach on a set of Twitter messages that are tagged with the location from which the message was sent. This dataset is more challenging than the geotagged Wikipedia articles, because Twitter texts are shorter, tend to contain unstructured text, and may not contain information about the location from where the message was sent in the first place. Nevertheless, we make some interesting observations about potential use of our geolocation algorithm for this type of input. We explain how we use the Spark framework for data analytics to collect and process our test data.
In general, classification-based approaches for location tagging may be reaching their upper limit for accuracy, but our precision-focused approach has high accuracy for some texts and shows significant potential for improvement overall.
\section{Introduction}
This paper explores the problem of extracting location information from text. The goal is to assign high precision geographical coordinates to places that are mentioned in texts from various sources.

This section discusses some background information and previous work in the area of location tagging. In the literature this is also referred to as geotagging or geolocation. Section \ref{ch:Algorithm} describes in detail the geolocation algorithm that we developed to approach this problem. We test this algorithm using geotagged Wikipedia articles in Section \ref{ch:Results}, and geotagged Twitter messages in Section \ref{ch:Results-Twitter}. Finally, we summarize the results and discuss some future work in Section \ref{ch:Conclusion}.
Appendix \ref{AppendixA} lists the external software that was used in the implementation of our project, Appendix \ref{AppendixB} describes how the Wikipedia test data for Section \ref{ch:Results} was obtained, and Appendix \ref{AppendixC} describes how the Twitter test data for Section \ref{ch:Results-Twitter} was composed.
\ref{AppendixD} provides links to where our code and data is available for download online.

\subsection{Previous Work using Classification Approaches}
Most studies in location tagging formulate it as a classification problem and use various machine learning approaches to solve it.
Classification problems begin with a defined set of classes. In location tagging, these classes can take many forms including cities, countries, or areas within a range of latitude and longitude coordinates. The goal of the classification problem is to assign a class (or a ranked list of most likely classes) to each text. Note that this problem is obviously most relevant for texts that indeed talk about a specific location, or a set of locations, but one can attempt to use the resulting algorithms to geotag any text.

For example, several papers in the location tagging literature \cite{Wing,Roller} use test data sets of Wikipedia articles about topics with a well-defined geographical location, and for which the Wikipedia authors have added a geographical location tag to each article. For these geotagged Wikipedia articles, the geographical coordinates in the (primary) location tag that was added by the authors is considered the single ``true location" of the text (but note that, clearly, the text may mention multiple locations). This data set can be used to train machine learning models and test their accuracy with respect to the ``true location" measure.

Wing and Baldridge created classes using simple geodesic grids of varying sizes \cite{Wing}. For example, one of their grids divided the surface of the Earth into 1\degree\ by 1\degree\ regions. Each of these regions was a class for their classification problem. They compared different models on various grid sizes, and tested these models using geotagged Wikipedia articles. They measured the distance between the article's true location and the location that was predicted by the model, and their best model achieved a median distance of 11.8 km.

Roller et al. realized that defining classes using a uniform latitude and longitude grid can be sensitive to the distribution of documents across the Earth \cite{Roller}. They looked at building an adaptive grid, which attempts to define classes such that each class has a similar number of training documents. They also tested their models using geotagged Wikipedia articles, and found a median error of 11.0 km. This is an improvement over the previous work \cite{Wing}.

Han et al. focused on location tagging of Twitter messages \cite{TwitterGeolocation}. They attempted to find the primary home location of a user (which is the city in which most of their tweets occur, and is known for their training/testing data set) by assembling all their tweets into a single document, which was then used as input to their models. Their classes were major cities along with their surrounding area. Their best model that only used the text of the tweet obtained a median error of 170 km. They obtained much greater accuracy when they incorporated additional information such as the home location that was listed on the user's profile, but this type of data is not available for general text geolocation.

One of the major challenges with Twitter messages is their unstructured nature. Tweets often contain spelling errors, abbreviations, and grammar mistakes which can make them difficult to interpret. Furthermore, some early work on our project discovered that most geotagged tweets (tweets that contain metadata with the GPS coordinates of the location the tweet was sent from) contain little geographic information in their text. This means that there is a very low threshold for the maximum accuracy we can expect to achieve when attempting to apply any geolocation algorithm to the text of tweets, attempting to establish where they were sent from only based on the text they contain. 

\subsection{Drawbacks of Classification Approaches}

It was thus decided that formulating our problem as a classification problem would not be feasible, as our goal was to obtain high precision. For example, if a text mentions the CN Tower in the city of Toronto then we want to return the coordinates of that building, rather than the coordinates for Toronto. Formulating this as a classification problem with this level of precision would ultimately require defining a class for every named location in the world. Furthermore, to apply these machine learning approaches we would need to have a large training set, ideally with multiple sample texts for each class. Obtaining this training data would be difficult, and even if the data was available it would likely be computationally infeasible to train such a model.
In general, we did not want to use any approach that relies heavily on training data, as we want our location tagger to be as general as possible. A tagger trained with Wikipedia articles might show worse performance when given other types of text, such as news articles or tweets. Acquiring good training data that is representative of all the types of texts we want to geolocate would be very difficult.
So we abandoned the classification approaches and turned towards natural language processing.

\subsection{Natural Language Processing}
\label{sec:NLP}

Natural language processing (NLP) encompasses the tasks that are related to understanding human language. In this paper we wish to understand location references in text, so it is natural to apply NLP techniques to this problem.

\paragraph{Part of Speech Tagging}

Part-of-speech (POS) tagging is the process of assigning part-of-speech tags to words in a text. The tagset may vary by language and the type of text, but typically includes tags such as nouns, verbs, and adjectives. Assigning these tags is an important step towards understanding text in many NLP tasks.

The Stanford Natural Language Processing Group provides a state-of-the-art POS tagger based on the research of Toutanova et al. \cite{StanfordPOS2000,StanfordPOS2003}. They use various models that observe features of words in the text (such as capitalization, for example) and predict the most likely part-of-speech tag for each word. They train these models using a large set of annotated articles from the Wall Street Journal.

Their software has two main types of models. One of these types is called the left-3-words model. The tag that is applied to each word in this model is influenced by the tags assigned to some of the previous words in the sentence.
The other type of model is known as the bidirectional model \cite{StanfordPOS2003}. Tags in these models are influenced by tags on both sides of the word, not just the previous words. This makes these models more complex. They have a slight accuracy improvement over the left-3-words models, but at the cost of an order of magnitude longer running time.
In this project we use Stanford's POS tagger with one of their pre-trained left-3-words models (see also Appendix \ref{AppendixA}). How our geolocation algorithm uses this software will be discussed in Section \ref{sec:LocationExtraction}.

\paragraph{Named Entity Recognition}

A Named Entity Recognizer (NER) finds proper nouns in text and determines to what type of entity the noun refers. This is a valuable tool in our approach to location tagging, as we are primarily looking for locations that are named entities in the text.
The Stanford Natural Language Processing Group also provides a NER based on the work of Finkel et al. \cite{StanfordNER}. Their software uses a conditional random field model to identify different classes of named entities in text. Their pre-trained model that was used in this paper attempts to identify each named entity as a location, person, or organization (see Appendix \ref{AppendixA}). In Section \ref{sec:LocationExtraction} our algorithm will use entities that are identified as locations by this software.

\paragraph{Named Entity Disambiguation}

After completing an NER task there is often some ambiguity. For example, if the NER software determines that some text mentions a location called \textit{London}, does it refer to a city in the UK or a city in Ontario? Choosing the correct interpretation is called disambiguation.

Habib \cite{Mena} explored both the extraction and disambiguation steps for named entity recognition, along with the link between these steps. He discovered that there is a feedback loop, where results from the disambiguation step can be used to improve the extraction step. This occurs because the disambiguation step can help to identify false positives, that is, words that were identified as a named entity by the extraction step but are not true named entities.
Named entities which refer to locations are called \textit{toponyms}, and a major portion of Habib's \cite{Mena} work discusses the extraction and disambiguation of toponyms in semi-formal text. He disambiguated toponymns in vacation property descriptions to determine in which country the property was located. While country-level locations do not provide the level of precision we desire for this project, his work served as inspiration for many steps of the algorithm we present in Section \ref{ch:Algorithm}.

More generally, named entity disambiguation has been an active topic of research throughout the past decade, see, e.g., \cite{milne2008learning,kulkarni2009collective,hoffart2011robust,usbeck2014agdistis,moro2014entity} and references therein.
For example, the AIDA system \cite{hoffart2011robust,yosef2011aida} is a general framework for collective disambiguation exploiting the prior probability of an entity being mentioned, similarity between the context of the mention and its candidates, and the coherence among candidate entities for all mentions together. AIDA uses knowledge bases such as DBpedia or YAGO and employs a graph-based algorithm to approximate the best joint mention-entity mapping. Compared to general named entity disambiguation frameworks such as AIDA, our contribution in this paper is that we present, as part of other new elements in our broader geolocation algorithm, a new named entity disambiguation approach that is specifically suited for disambiguating potential location references. It achieves high accuracy by relying on new distance-based scoring functions to measure the coherence between candidate locations. These distance-based scoring functions are used to disambiguate between multiple possible interpretations of word sequences and multiple possible locations for a potential location reference. Our approach, which uses the OpenStreetMap knowledge base \cite{haklay2008openstreetmap}, is specific to geolocation, and provides a high-precision and versatile method for geolocation of texts. 


\section{Methods}
\label{ch:Algorithm}

Given a segment of text, we want to find locations that are mentioned in the text. For example, in a text that states ``Bob drove from Waterloo to Toronto'' we want to find the names \textit{Waterloo} and \textit{Toronto} and determine which locations are meant by those words. Does \textit{Waterloo} mean a city in Ontario, Iowa, Belgium, or somewhere else? The mention of \textit{Toronto}, which can refer to another city in Ontario, suggests that the correct answer is Waterloo, Ontario. This type of reasoning was developed into a geolocation algorithm that is formally described in this section.

Section \ref{sec:AlgOverview} gives a high-level overview of the algorithm. Section \ref{sec:Terminology} defines the terminology we will use in the rest of the paper. Section \ref{sec:LocationExtraction} describes how names like \textit{Waterloo} and \textit{Toronto} are extracted from the text using part-of-speech tagging and named entity recognition. Section \ref{sec:Nominatim} describes how we discover that geographic names like \textit{Waterloo} can refer to multiple locations (e.g., Ontario, Iowa, or Belgium) by searching a knowledge base. Section \ref{sec:Disambiguation} describes how we disambiguate between the possible locations for each name (e.g., determine that \textit{Waterloo} and \textit{Toronto} refer to cities in Ontario) and between multiple interpretations for word sequences. Finally, the complete geolocation algorithm is summarized in Section \ref{sec:AlgSummary}.


\subsection{Overview}
\label{sec:AlgOverview}

Algorithm \ref{alg:SimpleOverview} gives a general high-level overview of the steps in the geolocation algorithm we propose in this paper.

\begin{algorithm}
\caption{Simplified overview of geolocation algorithm}
\label{alg:SimpleOverview}
\begin{algorithmic}[1]
\State Extract potential location references from the text. This is described in detail in Section \ref{sec:LocationExtraction}.
\State Search for each potential location reference in a knowledge base. This will give a list of result locations that are possible matches for the reference. This is described in Section \ref{sec:Nominatim}.
\State For each potential location reference, determine to which of the knowledge base results it most likely refers, and resolve conflicting interpretations of sequences of words. This is called disambiguation, and is described in detail in Section \ref{sec:Disambiguation}.
\end{algorithmic}
\end{algorithm}

Section \ref{sec:Terminology} will define some terminology which will allow us to write Algorithm \ref{alg:SimpleOverview} more precisely. Section \ref{sec:AlgSummary} will summarize Section \ref{ch:Algorithm} to give a more detailed version of Algorithm \ref{alg:SimpleOverview}.

\subsection{Terminology}
\label{sec:Terminology}

Before we continue with the description of our geolocation algorithm, we need to precisely define our terminology.

Consider, as an example, the following short text: ``Let's go shopping at Conestoga Mall in Waterloo. Waterloo lies between London and Guelph.'' This text talks about three cities in the Province of Ontario, Canada, and about a shopping mall in one of the three cities.

\begin{itemize}
\item A \textbf{term} is a word or sequence of adjacent words in the text. If a sequence of words occurs multiple times in the text, then every occurrence is considered a different term. Our algorithm will initially consider any sequence of words that occurs in the text as a potential location reference, i.e., it will initially consider all terms (sequences of words) of length one, two, three, etc.
\item Using part-of-speech tagging, named entity recognition, and a set of rules, the initial full set of terms (with their associated locations in the sentence) is reduced to a smaller set of terms that are judged potential location references. This (reduced) \textbf{term set} is called $T$. For the text example above, the term set $T$ would be \{Conestoga, Mall, Conestoga Mall, Waterloo, Waterloo, London, Guelph\} (with the location in the sentence also attached to each term; we omit to show this for simplicity). 
\item A word sequence may have multiple possible \textbf{interpretations}. For example, in the word sequence ``Conestoga Mall'', the two-word term ``Conestoga Mall'' itself may refer to a location, or, alternatively, the author of the text may have intended that either (or both) of the separate terms ``Conestoga'' or ``Mall'' refer to a location. To allow the disambiguation step to decide between these two mutually exclusive interpretations for the word sequence ``Conestoga Mall'', all three of the terms ``Conestoga'', ``Mall'', and ``Conestoga Mall'' are initially retained in the term set $T$. We say that, for the word sequence ``Conestoga Mall'', the term ``Conestoga Mall'' \textbf{conflicts} (or \textbf{overlaps}) with the terms ``Conestoga'' and ``Mall'', and the disambiguation step will choose one of the mutually exclusive interpretations by eliminating conflicting terms using an approach that employs distance-based scoring functions.
\item We say that each term corresponds to a \textbf{phrase}, where every phrase is only recorded once for the text. The \textbf{set of unique phrases} for the text is called $P$. Word sequences that occur multiple times in the text only occur once in $P$. For the text example above with term set $T$, the corresponding phrase set $P$ would be \{Conestoga, Mall, Conestoga Mall, Waterloo, London, Guelph\}. (The second occurrence of ``Waterloo'' is removed.) Note that $|P| \leq |T|$.
\item A \textbf{result} is a single geographical location that is listed in the knowledge base. For each phrase $p \in P$ we seek a set of results $R^p$ from the knowledge base. For any term $t \in T$, we let $t_{phr}$ be the phrase in $P$ associated with the term. So $R^{t_{phr}}$ is the set of results for the phrase corresponding to term $t$. As a shorthand, we also use $R^t$ instead of $R^{t_{phr}}$. When terms occur multiple times in a text, we will assume that they refer to the same potential location. 
For example, a text with two occurrences of \textit{Waterloo} will assume both terms refer to the same location. However, terms with phrases \textit{Waterloo Ontario} and \textit{Waterloo Belgium} will not be assumed to refer to the same location as they have different phrases.
Therefore, our disambiguation step will attempt to determine a single most likely location for each phrase $p \in P$, using an approach that employs distance-based scoring functions. 
\end{itemize}

We can now give an overview of the geolocation algorithm using this terminology: Algorithm \ref{alg:TechnicalOverview} is a re-writing of Algorithm \ref{alg:SimpleOverview} using the terms defined above.

\begin{algorithm}
\caption{Technical overview of geolocation algorithm}
\label{alg:TechnicalOverview}
\begin{algorithmic}[1]
\State Find all terms in the text that are potential location references. This is done using part-of-speech tagging and named entity recognition, and gives us the sets $T$ and $P$.
\State For each phrase $p \in P$, use $p$ as a search query in the knowledge base. The results of the query are the set $R^p$.
\State Reduce the set $T$ to remove terms which conflict with each other. Update the set $P$ accordingly to reflect changes in $T$. For each phrase $p \in P$, match $p$ to a single result $r \in R^p$. This is done by assigning distance-based scores to each result for every term, and selecting results with the greatest score in a greedy iterative fashion.
\end{algorithmic}
\end{algorithm}

\subsection{Location Extraction}
\label{sec:LocationExtraction}

The extraction phase of the algorithm uses the part-of-speech (POS) tagger and named entity recognizer (NER) from Stanford that were described in Section \ref{sec:NLP}. The entire text is tagged by both of these models. Based on the output of these taggers, the algorithm creates the set of terms $T$ which represents the potential location references in the text we consider further.

The development of our approach was initially guided by short unstructured texts coming from classified ads for rental housing, and we found that relying on NER tags only did not provide sufficient accuracy. Indeed, if the NER were perfect, we could simply use the NER to find all locations in the text and this step of the algorithm would be complete. Instead, our approach supplements the NER tags with the POS tagger, and considers more potential location references that the ones that are provided by the NER alone. The motivation for some of the heuristic algorithmic choices made in this section is further illustrated in an example at the end of the section.

\paragraph{Tagging the Text}
The POS tagger from Stanford assigns tags to each word in the text. The full list of possible tags is given in \cite{POSGuidelines}, but our algorithm only looks for a subset of tags which are relevant to our problem. Some of these tags are grouped together and considered to be equivalent by our algorithm. The POS tags used and how they are grouped are described in Table \ref{table:POSTags}.
The NER tags each word in the text with one of four possibilities: LOCATION, PERSON, ORGANIZATION, or O (meaning ``other''). 
When we initially extract relevant terms (see below), words with a LOCATION tag from the NER are considered equivalent to words with a noun tag from the POS tagger. However, when reducing the size of the term set $T$ in a subsequent step, we will give preference to words with LOCATION tags from the NER.

\begin{table}
\begin{tabular}{|c|c|c|}
\hline
\textbf{POS Tag} & \textbf{POS Tag Description} & \textbf{Our Grouping} \\
\hline
CC & Coordinating conjunction & conjunction \\
\hline
CD & Cardinal number & adjective \\
\hline
IN & Preposition or subordinating conjunction & preposition \\
\hline
JJ & Adjective & adjective \\
\hline
NN & Noun, singular or mass & noun \\
\hline
NNS & Noun, plural & noun \\
\hline
NNP & Proper noun, singular & noun \\
\hline
NNPS & Proper noun, plural & noun \\
\hline
TO & to & preposition \\
\hline
\end{tabular}
\caption{Part of Speech tags used in our geolocation algorithm.}
\label{table:POSTags}
\end{table}

\paragraph{Extracting Terms}

After each word in the text has been tagged by both the POS tagger and the NER, we build our set of terms $T$ which holds all potential location references. A word sequence in the text is considered a potential location reference if it satisfies two properties:
\begin{enumerate}
\item All words in the sequence must be tagged with a noun or adjective tag from Table \ref{table:POSTags}, or a LOCATION tag from the NER.
\item The sequence must contain at least one word with a noun tag from the POS tagger or a LOCATION tag from the NER (so sequences of only adjectives are not considered).
\end{enumerate}

The details about why these properties were chosen are described in the rest of this section. Note that for the remainder of this section, a \emph{noun} refers to any word that is tagged with LOCATION by the NER, or with one of the noun tags in Table \ref{table:POSTags} by the POS tagger (but not to adjectives).

Some locations, such as \textit{New York}, have multiple words in their name. Each word is tagged individually by the POS tagger and NER, so we need to consider these multi-word possibilities when building $T$.
If multiple nouns occur adjacent to each other in the text, then we do not know if these nouns refer to multiple locations or to one location. So we add all possibilities to the set $T$, and we will resolve conflicts in Section \ref{sec:Disambiguation}.
For example, if the algorithm discovers the phrase ``New York City'' in the text, with each of the three words tagged as a noun (note that the POS actually tags ``New" as a noun due to the capitalization), then the algorithm would add six terms to the set $T$: \textit{New}, \textit{York}, \textit{City}, \textit{New York}, \textit{York City}, and \textit{New York City}.

During the development of our geolocation algorithm it was discovered that some location references contain words tagged as adjectives. For example, given the text ``georgian college'' the POS tagger decides that \textit{georgian} is an adjective and that \textit{college} is a noun. The word \textit{georgian} is part of the name, even though it was not tagged as a noun. So our algorithm was modified to consider adjectives when they are part of a multi-word phrase with other nouns. However, it does not consider phrases that only contain adjectives (otherwise we could simply add adjectives to the noun group in Table \ref{table:POSTags}). In the ``georgian college'' example, both \textit{college} and \textit{georgian college} are added to $T$.
Also, if a text contains a street address, we want the street number to be part of the term. This will allow for greater precision in finding the proper location. Numbers are treated the same way as adjectives for this reason. So the example text ``200 University Avenue'' would generate five terms: \textit{University}, \textit{Avenue}, \textit{University Avenue}, \textit{200 University}, and \textit{200 University Avenue}.

Note that our set $T$ is not finalized yet. Some of the terms we added in this section may be removed for efficiency reasons, as described next.

\paragraph{Removing Terms}


The amount of time required for the knowledge base searches in Section \ref{sec:Nominatim} grows linearly with $|P|$, and the amount of time required for the disambiguation in Section \ref{sec:Disambiguation} is cubic in $|T|$. Furthermore, including too many terms in $T$ which are not true location references may make it difficult to disambiguate properly. Therefore it is advantageous to keep the size of $T$ small.

The first step in reducing $T$ is to check whether any of the terms contain words that were tagged as locations by the NER model. If there are LOCATION tags for any word in any term in $T$, then we keep only terms which contain at least one word tagged with LOCATION. All others are removed from the set $T$. The idea behind this is that if the NER finds words that are locations, then terms with these words are the best bet for successful disambiguation and geolocation.

If no words were tagged with LOCATION in the entire text, then the geolocation algorithm must rely solely on the results of the POS tagger. However, the set of nouns in a text can be quite large, so we still wish to filter $T$.

In the case where no LOCATION tags were found, the next step is to look for terms that occurred after prepositions. Many prepositions describe spatial relationships, so they can be strong indicators that a term does refer to a location. For example, the text ``Bob travelled from Waterloo'' contains the preposition \textit{from}. In this case, the preposition indicates that \textit{Waterloo} is a location.
However, prepositions can describe not only spatial relationships, but also temporal ones. For example, ``Bob lived there for five years'' uses \textit{for} as a temporal preposition. This type of preposition should not be included in our algorithm. The implementation that was used in this paper explicitly ignored the word \textit{for} as a preposition, as this was observed to increase accuracy for some development examples (from classified ads). All other prepositions identified by the POS tagger were retained. Future work could refine this list to exclude more non-spatial prepositions.

If the text contains terms that occur after prepositions but no words tagged with LOCATION, then these terms are retained while all others are discarded from $T$. A term is considered to be after a preposition if all words between the preposition and the term are tagged with any of the tags in Table \ref{table:POSTags}. It is for this reason that conjunctions are included in that table. Including conjunctions ensures that a text such as ``Guests travelled from Waterloo and Toronto'' will consider both Waterloo and Toronto.

Finally, if the text contains no LOCATION tags and there are no terms that follow prepositions, then no terms are removed from $T$.

\paragraph{Postal Codes}

Some of the development example texts contained postal codes. Postal codes can give very precise location information, so regular expressions were used to find postal codes that match the formats used by Canada, the United States, and the Netherlands. (The Netherlands were included because Spotzi, a partner company for this paper, has a branch there.) All occurrences of postal codes are added to the set $T$ if they are not already in the set due to previous steps. This occurs after the filtering to remove terms that was described above.

\paragraph{Example}

To demonstrate how the extraction step works, we will walk through an example in this section. This short example was used in the development of our geolocation algorithm, and shows many of the cases that were described in the previous sections. We will continue using this example in Section \ref{sec:Disambiguation}.

The text is from a Kijiji classifieds listing, and consists of a single sentence: ``A beautifull clean house for rent, Walking distance to RVH and Georgian college.'' This text is not properly structured. In particular, it has a spelling error and some improper capitalization. This makes the text challenging.

First we use the NER and POS tagger to tag all words in the text. The full list of tags for this example is given in Table \ref{table:POSExample}. We can see that the NER assigns no LOCATION tags in this text, so we must rely solely on the POS tags.

\begin{table}
\begin{tabular}{|c|c|c|c|}
\hline
\textbf{Word} & \textbf{NER Tag} & \textbf{POS Tag} & \textbf{POS Tag Description} \\
\hline
A & O & DT & Determiner \\
\hline
beautifull & O & NN & Noun, singular or mass \\
\hline
clean & O & JJ & Adjective \\
\hline
house & O & NN & Noun, singular or mass \\
\hline
for & O & IN & Preposition or subordinating conjunction \\
\hline
rent & O & NN & Noun, singular or mass \\
\hline
, & O & , &  \\
\hline
Walking & O & VBG & Verb, gerund or present participle \\
\hline
distance & O & NN & Noun, singular or mass \\
\hline
to & O & TO & to \\
\hline
RVH & O & NN & Noun, singular or mass \\
\hline
and & O & CC & Coordinating conjunction \\
\hline
Georgian & O & JJ & Adjective \\
\hline
college & O & NN & Noun, singular or mass \\
\hline
\end{tabular}
\caption{NER and POS tags for example text.}
\label{table:POSExample}
\end{table}

Next, we start building the set $T$. We take all nouns, along with adjectives that are adjacent to them. So our set $P$ that corresponds to $T$ is:
\begin{align*}
P = \{ & \textit{beautifull}, \textit{beautifull clean}, \textit{beautifull clean house}, \textit{clean house}, \textit{house}, \\
& \textit{rent}, \textit{distance}, \textit{RVH}, \textit{Georgian college}, \textit{college} \}
\end{align*}

Now we start to reduce this set. We have no terms that contain words tagged with LOCATION, but we do have terms that occur after prepositions.
The preposition \textit{to} occurs before \textit{RVH}, so the term for \textit{RVH} remains in $T$. Similarly, \textit{Georgian college} and \textit{college} are considered to occur after this preposition, because they are separated from this preposition only by tags in Table \ref{table:POSTags}.
The terms with phrases \textit{beautifull}, \textit{beautifull clean}, \textit{beautifull clean house}, \textit{clean house}, \textit{house}, and \textit{distance} are all discarded because they do not occur after a preposition. The term with \textit{rent} occurs after the preposition \textit{for}, but, as described in Section \ref{sec:LocationExtraction}, this is assumed to not be a spatial preposition, so \textit{rent} is also discarded.

So the set $P$ that corresponds to the final set $T$ is:
\begin{align*}
P = \{ & \textit{RVH}, \textit{Georgian college}, \textit{college} \}
\end{align*}
This set is used in the next step of the geolocation algorithm.

\subsection{Searching the Knowledge Base}
\label{sec:Nominatim}

After we generate the set $T$ (and the set $P$ that corresponds to it), the next step is to search for each $p \in P$ in a knowledge base. OpenStreetMap is used as the knowledge base for this paper \cite{haklay2008openstreetmap}. The OpenStreetMap data is queried using a tool called Nominatim (\url{https://nominatim.openstreetmap.org/}).

OpenStreetMap is a database that contains mapping data for the entire globe, including roadways, cities, and points of interest. This data is created and corrected by a large community of contributors, and is freely available.
Nominatim is an open-source tool that allows users to search the OpenStreetMap data. Using clever indexing and a significant amount of system resources, Nominatim allows queries of the vast OpenStreetMap data to be completed in seconds.\footnote{
There are donated servers running Nominatim which are free for light use, but they have a usage policy to prevent users from overloading the server. This usage policy gives a limit of one query per second. Our geolocation algorithm requires one query for each $p \in P$, so anyone who wants to use it extensively should set up their own Nominatim server. The tests in Section \ref{ch:Results} did use the freely available servers with a delay written into the code to ensure that at least one second passed between each query sent to Nominatim. This meant the geolocation algorithm completed much slower than it would otherwise, but it saved time on the engineering effort required to set up a Nominatim server.
}

Nominatim allows a number of parameters to be specified with the query. For example, searches can be limited to a particular region or country. One can also specify the maximum number of search results that should be returned for each query. In this paper the maximum number was set to 10. If Nominatim does not produce any result for a phrase, we discard all terms in $T$ that correspond to the phrase.
If the number of results $\geq 1$, then we use these results in the disambiguation step (step 3 of Algorithm \ref{alg:TechnicalOverview}), which selects for each phrase $p$ the correct result in $R^p$. Each result $r \in R^p$ contains latitude and longitude coordinates which are used to calculate distances between results as described in Section \ref{sec:Disambiguation}.

Another field that is included with each Nominatim search result is called ``importance''. This field is not well-documented, but after reading the source code it appears that a number of different factors are used to calculate this importance. These factors include:
\begin{itemize}
\item The type of the location (building, city, country, etc.)
\item The PageRank of the Wikipedia article about the location (if applicable)
\item The string similarity between the query string and the name of the location
\end{itemize}
We use the importance field as a tie-breaker in some steps of the disambiguation algorithm in Section \ref{sec:Disambiguation}. It is also used by Nominatim to sort results. If Nominatim finds more than 10 results for a query, it will return only the 10 most important results.

\paragraph{Example}
Here we continue the example from Section \ref{sec:LocationExtraction}. When we last saw this example we had three terms, and their corresponding phrases were
\[ P = \{ \textit{RVH}, \textit{Georgian college}, \textit{college} \} \]
We use each phrase as a search query with Nominatim. However, for this example we will limit Nominatim to return a maximum of three results for each phrase (instead of 10, which was used for the rest of this paper).
The results for each query are given in Table \ref{table:Nominatim:Example}. Nominatim found only one result for \textit{RVH}, two for \textit{Georgian college}, and three for \textit{college}. The phrase \textit{college} would have more results if we increased the result limit.

\begin{table}
\newcolumntype{Q}{>{\centering\arraybackslash} m{14mm} }
\newcolumntype{A}{>{\centering\arraybackslash} m{30mm} }
\newcolumntype{L}{>{\centering\arraybackslash} m{15mm} }
\newcolumntype{O}{>{\centering\arraybackslash} m{20mm} }
\newcolumntype{N}{>{\centering\arraybackslash} m{3mm} }
\begin{tabular}{|Q|N|Q|A|L|O|}
\hline
\multirow{2}{*}{\textbf{Query}} & \multicolumn{5}{|c|}{\textbf{Results}} \\
\cline{2-6}
& \textbf{\#} & \textbf{Name} & \textbf{Address} & \textbf{Latitude} & \textbf{Longitude} \\
\hline
RVH & 1 & Royal Victoria Regional Health Centre & 201, Georgian Drive, Barrie, Ontario, Canada & 44.41 & -79.66 \\
\hline
\multirow{2}{*}{\parbox{15mm}{\centering Georgian college}} & 1 & Georgian College & Georgian Drive, Barrie, Ontario, Canada & 44.41 & -79.67 \\
\cline{2-6}
& 2 & Georgian College &  Raglan Street, Collingwood, Ontario, L9Y3J4, Canada & 44.48 & -80.19 \\
\hline
\multirow{3}{*}{\parbox{15mm}{\centering college}} & 1 & College & Yonge Street, Church-Wellesley Village, Toronto, Ontario, M5B 2H4, Canada & 43.66 & -79.38 \\
\cline{2-6}
& 2 & College & Fairbanks North Star, Alaska, United States of America  & 64.86 & -147.80 \\
\cline{2-6}
& 3 & College & Los Ba\~nos, Laguna, Calabarzon, Philippines & 14.16 & 121.24 \\
\hline
\end{tabular}
\caption{Example nominatim results.}
\label{table:Nominatim:Example}
\end{table}

\subsection{Disambiguation}
\label{sec:Disambiguation}

In this section we describe in detail how our novel distance-based disambiguation procedure works.
We begin with an overview of the approach using two simple examples.

\paragraph{Overview of Distance-Based Disambiguation Approach}

Consider again the short text ``Let's go shopping at Conestoga Mall in Waterloo. Waterloo lies between London and Guelph.'' The term set $T$=\{Conestoga, Mall, Conestoga Mall, Waterloo, Waterloo, London, Guelph\}, and the phrase set $P$=\{Conestoga, Mall, Conestoga Mall, Waterloo, London, Guelph\}.

The disambiguation phase of our geolocation algorithm has two goals. First, it serves to decide between mutually exclusive interpretations of word sequences (i.e., it decides which terms to retain when terms are conflicting (or overlapping), see Section \ref{sec:Terminology}). In the example, the sequence ``Conestoga Mall'' may refer to a single location called ``Conestoga Mall'', or, alternatively, one (or both) of the terms ``Conestoga" and ``Mall'' may refer to a location instead. The disambiguation step will select one of these conflicting interpretations. Second, for each phrase $p \in P$, we will use the disambiguation step to select one of the results $r$ in the result set $R^p$. Both of these processes will rely on the physical distances between result locations, and will be based on a table of pairwise distances that is computed between the different candidate results for the different terms in the text, as, for example, in Tables \ref{table:scoring2} and \ref{table:scoring3}.

To calculate the distance between two results $a$ and $b$ from Nominatim, let $a_x$ and $b_x$ be their longitude coordinates, and let $a_y$ and $b_y$ be their latitude coordinates. 
The distance between $a$ and $b$ is then calculated by
\begin{equation}\label{eq:Distance}
d(a,b) = 6371 \arccos \left( \sin(a_y) \sin(b_y) + \cos(a_x) \cos(b_x) \cos(a_x - b_x) \right)
\end{equation}
Equation (\ref{eq:Distance}) is the great circle distance on a sphere, where 6371 is the mean radius of the Earth in kilometres. 

Let us first consider how disambiguation would work for the short text ``Waterloo lies between London and Guelph". For this sentence, there are no conflicting interpretations of word sequences, and we only have to resolve the multiple results for each phrase. Table \ref{table:scoring2} shows mutual distances between pairs of results for each of the terms in the term set $T$=\{Waterloo, London, Guelph\}. The first column of the table shows that the first result for Waterloo (the Ontario result) is located within a short distance from the Ontario results for London and Guelph (and, of course, the fourth column shows that the Ontario results for London and Guelph are also closeby). This indicates that the Ontario results for Waterloo, London and Guelph are the coherent choice for disambiguation. In this small example with three terms who have two results each, this is easy to see, but for larger problems with tens or more of terms and 5 or more results per term, we need a systematic procedure to find coherent groups of results. We use the following systematic approach.

\begin{table}
\newcolumntype{C}{>{\centering\arraybackslash} m{9mm} }
\begin{tabular}{|c|c|c|c|c|c|c|c|}
\cline{3-8}
\multicolumn{2}{c|}{} & \multicolumn{2}{|c|}{\textit{Waterloo}} & \multicolumn{2}{|c|}{\textit{London}} & \multicolumn{2}{|c|}{\textit{Guelph}} \\
\cline{3-8}
\multicolumn{2}{c|}{} & 1 & 2 & 1 & 2 & 1 & 2 \\
\hline
\multirow{2}{*}{\parbox{18mm}{\centering \textit{Waterloo}}} & 1 (Ontario) & x & x & 5797 & 79 & 24 & 1424 \\
\cline{2-8}
& 2 (Belgium) & x & x & 328 & 6198 & 6096 & 6956 \\
\hline
\multirow{2}{*}{\parbox{18mm}{\centering \textit{London}}} & 1 (UK) & 5797 & 328 & x & x & 5774 & 6655 \\
\cline{2-8}
& 2 (Ontario) & 79 & 6198 & x & x & 102 & 1386 \\
\hline
\multirow{2}{*}{\parbox{18mm}{\centering \textit{Guelph}}} & 1 (Ontario) & 24 & 6096 & 5774 & 102 & x & x \\
\cline{2-8}
& 2 (North Dakota) & 1424 & 6956 & 6655 & 1386 & x & x \\
\hline
\hline
\multicolumn{2}{|c|}{\textbf{Score $S_r^t$}} & -103 & -6424 & -6102 & -181 & -126 & -2810 \\
\hline
\end{tabular}
\caption{First example of distance-based disambiguation, for the sentence ``Waterloo lies between London and Guelph''. The Total Distance scoring function is used.}
\label{table:scoring2}
\end{table}

First we compute, for each result $r$ and every term $t$ (except the term for which $r$ is a result), the shortest distance $c(r,t)$ between result $r$ and any of the results for term $t$. For example, for the first column of Table \ref{table:scoring2}, $c(Waterloo (Ontario), London)$=79 and $c(Waterloo (Ontario), Guelph)$=24.
Then we compute, for every result $r$ of every term $t$, the scoring function $S_r^t$, where the value of the scoring function measures how good a candidate a result is as a choice for disambiguation.
The simplest scoring function we consider is the \textbf{Total Distance} scoring function, given here by $S^{t}_{r} = - \sum_{\substack{t_2 \in T, t_2 \ne t}} c(r, t_2)$. This scoring function is computed in the bottom line of Table \ref{table:scoring2}, for each of the six results. For example, in the first column of the table, -103 is the Total Distance score for result Waterloo (Ontario) of term Waterloo. It is the negative of the sum of the shortest distances of result Waterloo (Ontario) to the terms London and Guelph.
Our approach then proceeds by selecting the result with the overall highest score (closest to 0, i.e., smallest sum of distances) as the first disambiguation result. In our example, this means that result Waterloo (Ontario) is chosen for term Waterloo.
All results for Waterloo are then removed from the table (except for the result that was chosen), and the process repeats: the scoring functions of the reduced table are recomputed, and the result with the next largest score is selected to disambiguate the next term.
In this manner, the Ontario results for Waterloo, London and Guelph are selected. In practice, the process is more complex (see below), but this is a simple illustration of the general principle.

We next consider disambiguating the short text ``Let's go shopping at Conestoga Mall in Waterloo."
This is more complicated, because the word sequence ``Conestoga Mall" has two conflicting interpretations.
Table \ref{table:scoring3} shows mutual distances between pairs of results for each of the terms in the term set
$T$=\{Waterloo, Conestoga Mall, Conestoga, Mall\}. The term ``Conestoga Mall" conflicts (or overlaps) with
``Conestoga" and ``Mall" (since the author cannot have intended both as locations at the same time),
so no mutual distances are computed. The shortest distances $c(r,t)$ are computed as before.
When computing the scoring functions $S_r^t$ for every result $r$ of every term $t$, we now
introduce weights $W^{t_1}_{t_2}$ for any pair of terms $t_1,t_2 \in T$.
These weights are used to reduce bias in calculating scores when there are terms in
$T$ that conflict with each other and multiple mutually exclusive interpretations need to be considered.

For example, for the first column of Table \ref{table:scoring3}, when computing the scoring function
$S_r^t$ for result Waterloo (Ontario) of term Waterloo, we cannot just add the shortest distances of
Waterloo (Ontario) to all of the terms ``Conestoga Mall",  ``Conestoga" and ``Mall", as this would
bias the score since only one or at most two distances should be counted, given the mutually exclusive
interpretations of the word sequence ``Conestoga Mall". Since the correct interpretation has not been
chosen yet, the best we can do is to take all the three shortest distances into account, but we take a weighted
average of these three distances to reduce the bias in the procedure (compared to, for example, results for
term ``Conestoga Mall", which only have one distance in their scoring function). When computing the
scoring functions for the results of term $t_1$ (e.g., ``Waterloo"), we define weights $W^{t_1}_{t_2}$ with
respect to $t_1$ for all the terms $t_2$ that correspond to a word sequence with conflicting interpretations
(e.g., ``Conestoga Mall"), in such a way that the sum of the weights equals 1, and with equal weight for
every interpretation. Each of the two interpretations of the phrase``Conestoga Mall" is given an equal
weight of 1/2. Within the second interpretation (``Conestoga'' and ``Mall"), each of the terms have equal weight, so the final weights
are $W^{t_1}_{t_2}=1/4$ for $t_2=$``Conestoga", $W^{t_1}_{t_2}=1/4$ for $t_2=$``Mall", and
$W^{t_1}_{t_2}=1/2$ for $t_2=$``Conestoga Mall".
So for our example, for the first column of Table \ref{table:scoring3}, we compute
the scoring function $S_r^t$ for result Waterloo (Ontario) of term Waterloo using the \textbf{Weighted Distance}
scoring function, given here by $S^{t}_{r} = - \sum_{\substack{t_2 \in T, t_2 \ne t}} W^{t}_{t_2} c(r, t_2)$,
resulting in $S^{t}_{r}=-279.5$ for result Waterloo (Ontario) of term Waterloo.

\begin{sidewaystable}
\centering
\newcolumntype{C}{>{\centering\arraybackslash} m{9mm} }
\begin{tabular}{|c|c|c|c|c|c|c|c|c|c|}
\cline{3-10}
\multicolumn{2}{c|}{} & \multicolumn{2}{|c|}{\textit{Waterloo}} & \multicolumn{2}{|c|}{\textit{Conestoga Mall}} & \multicolumn{2}{|c|}{\textit{Conestoga}} & \multicolumn{2}{|c|}{\textit{Mall}} \\
\cline{3-10}
\multicolumn{2}{c|}{} & 1 & 2 & 1 & 2 & 1 & 2 & 1 & 2 \\
\hline
\multirow{2}{*}{\parbox{18mm}{\centering \textit{Waterloo}}} & 1 (Ontario) & x & x & 4 & 1495 & 523 & 3264 & 587 & 5797 \\
\cline{2-10}
& 2 (Belgium) & x & x & 6117 & 7376 & 6105 & 8974 & 6225 & 328 \\
\hline
\multirow{2}{*}{\parbox{18mm}{\centering \textit{Conestoga Mall}}} & 1 (Waterloo) & 4 & 6117 & x & x & x & x & x & x \\
\cline{2-10}
& 2 (Nebraska) & 1495 & 7376 & x & x & x & x & x & x \\
\hline
\multirow{2}{*}{\parbox{18mm}{\centering \textit{Conestoga}}} & 1 (Pennsylvania) & 523 & 6105 & x & x & x & x & 130 & 5778 \\
\cline{2-10}
& 2 (California) & 3264 & 8974 & x & x & x & x & 3541 & 8681  \\
\hline
\multirow{2}{*}{\parbox{18mm}{\centering \textit{Mall}}} & 1 (Washington DC) & 587 & 6225 & x & x & 130 & 3541 & x & x \\
\cline{2-10}
& 2 (London UK) & 5797 & 328 & x & x & 5778 & 8681 & x & x \\
\hline
\hline
\multicolumn{2}{|c|}{\textbf{Score}} & -279.5 & -4666.75 & -4 & -1495 & -653 & -6805 & -717 & -6106 \\
\hline
\end{tabular}
\captionof{table}{Second example of distance-based disambiguation, for the sentence ``Conestoga Mall in Waterloo".
There are two mutually exclusive interpretations for the word sequence ``Conestoga Mall". In the first interpretation, the combined term ``Conestoga Mall" refers to a potential location, and in the second interpretation the two separate terms ``Conestoga" and ``Mall" are considered potential locations. 
When determining the distance-based scoring function for each of the possible results for Waterloo, the distances to the terms ``Conestoga Mall", ``Conestoga" and ``Mall" are required, but since the two interpretations are mutually exclusive, weights are used in the Weighted Distance scoring function that is employed in the table.}
\label{table:scoring3}
\end{sidewaystable}

After the scoring functions are computed for all terms of each result in this way, we decide
on the first disambiguation result in the same greedy fashion as before:
the highest score in the bottom row of Table \ref{table:scoring3} is the -4 for result 1 of term
``Conestoga Mall". This result is selected for this term, and (in our simplest algorithm)
this settles at the same time the interpretation for the word sequence ``Conestoga Mall".
All results for the term ``Conestoga Mall", except for the one we choose,
are removed from the table, as well as the terms ``Conestoga"
and ``Mall" that conflict with it, and this finalizes the disambiguation since the Ontario location
for Waterloo is the one that is closest to the chosen result for ``Conestoga Mall".
In practice, the weighting procedure can be significantly more complex and may have
to be applied in a modified way when longer sequences of words may have a larger
number of possible interpretations (see below), but this example is a simple illustration
of the general principle.

Note that, in our general procedure, if a sequence of words with multiple
interpretations occurs several times in a text, then the interpretation is determined for
each occurrence separately (based on eliminating overlapping terms). 
This approach was chosen because the location of terms in the text
(overlap of adjacent terms) is important when disambiguating mutually exclusive interpretations.
In contrast, if a phrase occurs several times in a text, a single location result is selected
for all occurrences of the phrase. This simplifies the implementation.


In what follows, we provide complete details of the full distance-based disambiguation procedure.
Multiple scoring functions were tested, and they are described after giving full details on the weights
mechanism. Finally, two algorithms will be given that use the scores to perform the disambiguation.


\paragraph{Weights}
Weights are used when there are terms in the set $T$ that conflict with each other.
Terms conflict when they overlap in the text. This is best explained in detail with the
New York City example that was introduced in Section \ref{sec:LocationExtraction}.
For this example,  the extraction step finds the nouns \textit{New}, \textit{York}, and \textit{City} adjacent to each other in the text, and six terms are added to the set $T$: \textit{New}, \textit{York}, \textit{City}, \textit{New York}, \textit{York City}, and \textit{New York City}. The terms \textit{York} and \textit{New York} conflict with each other because the text could not have meant to refer to a location called \textit{York} and a different location called \textit{New York}. This is because the terms \textit{York} and \textit{New York} overlap. For this example, the full list of terms that conflict with any given term is provided
in Table \ref{table:TermConflicts}.
The purpose of the weights we will define in this section is to properly account for conflicting terms in the disambiguation step. 
Assigning weights to conflicting terms will allow us to reduce bias in the scoring functions until we can determine which interpretation of the text segment is correct.

\begin{table}
\begin{tabular}{|c|c|}
\hline
\textbf{Term} & \textbf{Terms that Conflict with it} \\
\hline
\textit{New} & \textit{New York}, \textit{New York City} \\
\hline
\textit{York} & \textit{New York}, \textit{York City}, \textit{New York City} \\
\hline
\textit{City} & \textit{York City}, \textit{New York City} \\
\hline
\textit{New York} & \textit{New}, \textit{York}, \textit{York City}, \textit{New York City} \\
\hline
\textit{York City} & \textit{York}, \textit{City}, \textit{New York}, \textit{New York City} \\
\hline
\textit{New York City} & \textit{New}, \textit{York}, \textit{City}, \textit{New York}, \textit{York City} \\
\hline
\end{tabular}
\caption{Conflicting terms for the word sequence ``New York City''.}
\label{table:TermConflicts}
\end{table}

We can think of the weight $W^{t_1}_{t_2}$ as the probability that term $t_2$ is a true location reference in the text given that $t_1$ is a true location reference. 
Weights are defined such that $0 \leq W^{t_1}_{t_2} \leq 1 \text{ } \forall t_1,t_2 \in T$. If no terms in $T$ conflict with each other then all weights are equal to 1. The remainder of this section describes how weights are calculated when some terms do conflict.

We begin our definition of weight with the case of two terms that conflict (i.e., overlap in the text):
\begin{equation}\label{eq:Weight:Conflict}
W^{t_1}_{t_2} = 0 \quad \forall t_1,t_2 \in T, t_1\ne t_2 \ | \ t_1 \text{ and } t_2 \text{ conflict}
\end{equation}
This simply expresses that conflicting terms cannot both refer to true locations at the same time, and it means
that distances between conflicting terms never need to be taken into account in scoring functions.
Thinking of the weights as the probabilities we described earlier, we can define:
\begin{equation}\label{eq:Weight:SameTerm}
W^{t}_{t} = 1 \quad \forall t \in T.
\end{equation}

Before we define the weight in the other cases, we need another definition. Conflicting terms appear in \textbf{groups}.
We define $G(t)$ to be the group of term $t$. $G(t)$ is the smallest set of all terms that overlap with $t$ or any of the other terms in $G(t)$. I.e., a group is a minimal set of terms such that all terms that overlap with any of the terms in the group are in the group. Note that $G(t)=\{t\}$ iff $t$ does not overlap with any other term.
Using this definition, we can conclude that all terms in Table \ref{table:TermConflicts} are in the same group.
Groups are disjoint, and each group corresponds to a segment (sequence of words) in the text. These segments are disjoint.

Groups with more than one term have multiple interpretations.
An interpretation of a group is a maximal subset of the group which contains terms that do not conflict.
An interpretation does not have to cover all words in the segment of text, but it must contain enough terms such that no non-conflicting terms can be added to the group.\footnote{In our New York City example all our interpretations cover all words in that segment of text. However, if the term \textit{New} were tagged as an adjective and thus missing from $T$, then $\{ \textit{York}, \textit{City} \}$ and $\{ \textit{New York}, \textit{City} \}$ would both be valid interpretations. However, $\{ \textit{York} \}$ would not be valid, since the term for \textit{City} can still be added.} Table \ref{table:Weight:Interpretations} lists all four interpretations for the New York City example.

When computing distance-based scores for the candidate results of a term $t_1$, we will, for every group $G\ne G(t_1)$, define weights $W^{t_1}_{t_2}$ with respect to $t_1$ for all the terms $t_2$ in group $G$. For every group $G$ these weights sum up to 1. If $G$ has only one element $t_2$ (which does not conflict with $t_1$ since $G\ne G(t_1)$), then $W^{t_1}_{t_2}=1$.
If $G$ has more than one element, the weights $W^{t_1}_{t_2}$ are defined as follows.
Let $q$ be the number of interpretations of group $G$, each with weight $1/q$, and let $n_i$ be the number of terms in interpretation $i$, each with weight $1/n_i$ in interpretation $i$.
Then
\begin{equation}\label{eq:Weight:othergroup}
W^{t_1}_{t_2} = \sum_{\substack{ \text{interpretations } i\\
 \text{ that contain } t_2}}  \frac{1}{q} \ \frac{1}{n_i}.
\end{equation}
%
For our New York City example, the weights given to each term in each interpretation are included in the last column of Table \ref{table:Weight:Interpretations}.
Table \ref{table:Weights:ExampleWeights} continues the calculations from Table \ref{table:Weight:Interpretations} to give the weights $W^{t_1}_{t_2}$ for the case where $t_2$ is a term in the New York City group and $t_1 \notin G(t_2)$. For example, this table tells us that when $t_1 \notin G(\textit{New})$ then $W^{t_1}_{\textit{New}} = \frac{5}{24}$.

\begin{table}
\newcolumntype{C}{>{\centering\arraybackslash} m{2.8cm} } 
\begin{tabular}{|C|C|C|C|}
\hline
\textbf{Interpretation Number} & \textbf{Terms In Interpretation} & \textbf{\# of Terms} & \textbf{Weight Given To Each Term} \\
\hline
1 & \textit{New}, \textit{York}, \textit{City} & 3 & $\frac{1}{4} * \frac{1}{3} = \frac{1}{12}$ \\
\hline
2 & \textit{New York}, \textit{City} & 2 & $\frac{1}{4} * \frac{1}{2} = \frac{1}{8}$ \\
\hline
3 & \textit{New}, \textit{York City} & 2 & $\frac{1}{4} * \frac{1}{2} = \frac{1}{8}$ \\
\hline
4 & \textit{New York City} & 1 & $\frac{1}{4} * \frac{1}{1} = \frac{1}{4}$ \\
\hline
\end{tabular}
\caption{Interpretations for ``New York City''.}
\label{table:Weight:Interpretations}
\end{table}

\begin{table}
\newcolumntype{C}{>{\centering\arraybackslash} m{1.7cm} } 
\begin{tabular}{|c|C|C|C|C|C|}
\hline
\textbf{Term} $t_2$ & \textbf{Weight from Interp. 1} & \textbf{Weight from Interp. 2} & \textbf{Weight from Interp. 3} & \textbf{Weight from Interp. 4} & \textbf{Weight} \\
\hline
\textit{New} & 1/12 & $0$ & 1/8 & $0$ & 5/24 \\
\hline
\textit{York} & 1/12 & $0$ & $0$ & $0$ & 1/12 \\
\hline
\textit{City} & 1/12 & 1/8 & 0 & 0 & 5/24 \\
\hline
\textit{New York} & $0$ & 1/8 & $0$ & $0$ & 1/8 \\
\hline
\textit{York City} & $0$ & $0$ & 1/8 & $0$ & 1/8 \\
\hline
\textit{New York City} & $0$ & $0$ & $0$ & 1/4 & 1/4 \\
\hline
\textbf{Interp. Total} & 1/4 & 1/4 & 1/4 & 1/4 &  \\
\hline
\end{tabular}
\caption{Weights $W^{t_1}_{t_2}$ for terms $t_2$ in the group corresponding to the segment ``New York City'' where $t_1 \notin G(t_2)$.}
\label{table:Weights:ExampleWeights}
\end{table}


Finally, we must define $W^{t_1}_{t_2}$ for the case where $t_1 \in G(t_2)$ and $t_1$ and $t_2$ do not conflict. As discussed earlier, the weight $W^{t_1}_{t_2}$ is the weight of $t_2$ when we assume that $t_1$ is in the text. Therefore, all we need to do to calculate the weight in this case is temporarily remove terms from $T$ that conflict with $t_1$. This will change the groups $G(t_1)$ and $G(t_2)$ such that $t_1$ is no longer in $G(t_2)$ (because $t_1$ no longer conflicts with any terms, $G(t_1) = \{ t_1 \}$). Our calculation of the weights then proceeds as before with Equation (\ref{eq:Weight:othergroup}).

For example, suppose we want to calculate $W^{\textit{New}}_{t_2}$ for $t_2 \in G(\textit{New})$. $W^{\textit{New}}_{\textit{New York}} = W^{\textit{New}}_{\textit{New York City}} = 0$ because those terms conflict. We remove those terms, so \textit{New} becomes the only member of its group. 
The remaining terms have only two interpretations: $\{\textit{York}, \textit{City}\}$ and $\{\textit{York City}\}$. When we calculate those weights we find that $W^{\textit{New}}_{\textit{York}} = W^{\textit{New}}_{\textit{City}} = \frac{1}{4}$ and $W^{\textit{New}}_{\textit{York City}} = \frac{1}{2}$.

\paragraph{Scoring Functions}
We now formulate eight different candidate scoring functions that can be used in the disambiguation step of
our geolocation algorithm. The accuracy obtained with these different scoring functions is tested in Section
\ref{ch:Results}. Note that all scoring functions are defined such that greater values of the scoring function indicate a ``better'' (higher) score.

The first scoring function we consider is called \textbf{Total Distance}:
\begin{equation}\label{eq:TotalDistance}
S^{t_1}_{r_1} = - \sum_{\substack{t_2 \in T. \\
                               W^{t_1}_{t_2} \neq 0}} c(r_1, t_2)
\end{equation}
For a term $t_1 \in T$ and a result $r_1 \in R^{t_1}$, this function simply adds up the minimum distance between $r_1$ and the closest result for each other term in $T$. Note that for any term $t$ such that $t_{phr} = t_{1,phr}$ (including the original term $t = t_1$) we define $c(r_1,t) = 0$. Therefore we do not need to explicitly ensure that $t_2 \neq t_1$ in the summation in Equation (\ref{eq:TotalDistance}). However, we do explicitly ensure that $W^{t_1}_{t_2} \neq 0$ so we do not consider conflicting terms. The Total Distance score is always negative, and values close to 0 indicate desirable scores (small sum of minimal distances).


The second scoring function we consider is called \textbf{Weighted Distance}:
\begin{equation}\label{eq:WeightedDistance} S^{t_1}_{r_1} = - \sum_{t_2 \in T} W^{t_1}_{t_2} c(r_1, t_2).
\end{equation}
This is equivalent to Equation (\ref{eq:TotalDistance}) except for the multiplication by the weight. The reasoning behind these weights and how they are calculated was described above.
Note that the minus sign was added to scoring functions (\ref{eq:TotalDistance}) and (\ref{eq:WeightedDistance}) 
to ensure that greater values of the scoring function indicate a ``better'' (higher) score (closer to zero, for these scoring functions).

Scoring functions (\ref{eq:TotalDistance}) and (\ref{eq:WeightedDistance}) may be sensitive to terms that are extremely far away from the others: in the sums (\ref{eq:TotalDistance}) and (\ref{eq:WeightedDistance}), one very large smallest distance may drown out the discriminative power of smaller smallest distances.
This can occur, for example, when the location extraction step (step 1 of Algorithm \ref{alg:TechnicalOverview}) extracts a phrase that is not intended to refer to a location in the text, but does have results in the knowledge base. The location extraction step is not perfect, so this is a common occurrence. The remaining scoring functions attempt to deal with this difficulty.

The next two scoring functions we consider are called \textbf{Inverse}:
\begin{equation}\label{eq:Inverse}
S^{t_1}_{r_1} = \sum_{\substack{t_2 \in T \\
                               t_{1,phr} \neq t_{2,phr} \\
                               W^{t_1}_{t_2} \neq 0}} \frac{1}{\max \left( c(r_1, t_2), 10^{-3} \right) },
\end{equation}
and \textbf{Weighted Inverse}:
\begin{equation}\label{eq:WeightedInverse}
S^{t_1}_{r_1} = \sum_{\substack{t_2 \in T \\
                               t_{1,phr} \neq t_{2,phr}}} \frac{W^{t_1}_{t_2}}{\max \left( c(r_1, t_2), 10^{-3} \right) }.
\end{equation}
They are similar to Equations (\ref{eq:TotalDistance}) and (\ref{eq:WeightedDistance}) respectively, except that we use the reciprocal of the $c(r,t)$ function. To avoid issues with division by zero, we take the maximum of $c$ and $10^{-3}$.
Due to the use of the reciprocal, small smallest distances increase the score strongly (as desired), and large smallest distances due to outliers have little effect.

Next we consider some normalization of Equation (\ref{eq:WeightedInverse}) to attempt to make scores more comparable to each other. Indeed, for a given text, the number of terms in the sums of Equation (\ref{eq:WeightedInverse}) can be quite different for different terms $t_1$, and the weights do not fully compensate for this. This may bias the scores. For example, for the example of Table \ref{table:scoring3}, the $S_{r_1}^{t_1}$ scoring function of Equation (\ref{eq:WeightedInverse}) has only one term for $t_1=$``Conestoga Mall'', but has two terms for $t_1=$``Conestoga'' or $t_1=$``Mall''. To remedy this difficulty, we consider the \textbf{Weighted Normalized Inverse} scoring function, which is given by
\begin{equation}\label{eq:WeightedNormalizedInverse}
S^{t_1}_{r_1} = \frac{ 
                                 \sum_{\substack{t_2 \in T \\
                                                            t_{1,phr} \neq t_{2,phr} }} 
                                 W^{t_1}_{t_2} 
                                 \left( \frac{\min_{r_1' \in R^{t_1} } \max \left(c(r_1', t_2), 10^{-3} \right) }
                                              {\max \left( c(r_1,t_2), 10^{-3} \right) } \right) 
                                }
                               { \sum_{\substack{t_2 \in T \\
                                                             t_{1,phr} \neq t_{2,phr} }} 
                                                             W^{t_1}_{t_2}
                               }.
\end{equation}
Equation (\ref{eq:WeightedNormalizedInverse}) was constructed by multiplying the numerator in Equation (\ref{eq:WeightedInverse}) by the minimum distance between any result for $t_1$ and any result for $t_2$, not just between result $r_1$ and any result for $t_2$. Finally, we divide the whole expression by the total weight of all terms we looked at. This means that the score $S^t_r$ is always between 0 and 1. A score of $S^t_r = 1$ means that when we consider the closest pair of results between $t$ and another term, $r$ is always the result from $t$ that is part of that closest pair.

In scoring functions (\ref{eq:Inverse}), (\ref{eq:WeightedInverse}), and (\ref{eq:WeightedNormalizedInverse}) we skip all terms $t_2 \in T$ that have the same phrase as $t_1$. However, if a phrase appears many times in the text then it is likely more important. So it might be helpful to give a bonus to the scores based on how often the phrase occurs. We add this feature to Equations (\ref{eq:Inverse}), (\ref{eq:WeightedInverse}), and (\ref{eq:WeightedNormalizedInverse}) to produce scoring functions \textbf{Inverse Frequency}:
\begin{equation}\label{eq:InverseFrequency}
S^{t_1}_{r_1} = \left( \sum_{\substack{t_2 \in T \\
                               t_{1,phr} \neq t_{2,phr} \\
                               W^{t_1}_{t_2} \neq 0}} \frac{1}{\max \left( c(r_1, t_2), 10^{-3} \right) } \right)  \sum_{\substack{t_2 \in T \\
                               t_{1,phr} = t_{2,phr}}} 1,
\end{equation}
\textbf{Weighted Inverse Frequency}:
\begin{equation}\label{eq:WeightedInverseFrequency}
S^{t_1}_{r_1} = \left( \sum_{\substack{t_2 \in T \\
                               t_{1,phr} \neq t_{2,phr}}} \frac{W^{t_1}_{t_2}}{\max \left( c(r_1, t_2), 10^{-3} \right) } \right) \sum_{\substack{t_2 \in T \\
                               t_{1,phr} = t_{2,phr}}} W^{t_1}_{t_2},
\end{equation}
and \textbf{Weighted Normalized Inverse Frequency}:
\begin{equation}\label{eq:WeightedNormalizedInverseFrequency}
S^{t_1}_{r_1} = \frac{ \sum_{\substack{t_2 \in T \\
                               t_{1,phr} \neq t_{2,phr}}} W^{t_1}_{t_2} \left( \frac{ \min_{r_1' \in R^{t_1} } c(r_1', t_2)}{\max \left( c(r_1,t_2), 10^{-3} \right) } \right) }{\sum_{\substack{t_2 \in T \\
                 t_{1,p} \neq t_{2,p}}} W^{t_1}_{t_2}} \sum_{\substack{t_2 \in T \\
                               t_{1,phr} = t_{2,phr}}} W^{t_1}_{t_2}.
\end{equation}
In Equations (\ref{eq:WeightedInverseFrequency}) and (\ref{eq:WeightedNormalizedInverseFrequency}) this is done by multiplying the original expression by the total weight of all terms with the same phrase. Since scoring function (\ref{eq:Inverse}) is the same as (\ref{eq:WeightedInverse}) but with all non-zero weights set to 1, we similarly set all non-zero weights to 1 to obtain the multiplicative factor in Equation (\ref{eq:InverseFrequency}).

We have now created eight different scoring functions, which we will test in Section \ref{ch:Results}. 
We will find that many of these scoring functions perform similarly. The Weighted Inverse Frequency scoring function (Equation \ref{eq:WeightedInverseFrequency}) will turn out to perform best in our Wikipedia tests.
Combined with the two disambiguation algorithms discussed in the next section, the eight scoring functions give 16 versions of our geolocation algorithm that will be compared. 

\paragraph{Disambiguation Algorithms}

We present two versions of the disambiguation step of our geolocation algorithm. The first version, called the 1-phase disambiguation algorithm, is described in Algorithm \ref{alg:1Phase}.

\begin{algorithm}
\caption{1-Phase disambiguation algorithm}
\label{alg:1Phase}
\begin{algorithmic}[1]
\State Calculate $W^{t_1}_{t_2} \text{ } \forall t_1,t_2 \in T$ using Equations (\ref{eq:Weight:Conflict}), (\ref{eq:Weight:SameTerm}), and (\ref{eq:Weight:othergroup}) \label{alg:1Phase:CalculateWeight}
\While{$(\exists p \in P | (|R^p| > 1))$ or $(\exists t_1,t_2 \in T | (W^{t_1}_{t_2} \neq 1 ))$} \label{alg:1Phase:while}
  \ForAll{$t \in T$}
    \ForAll{$r \in R^t$}
      \State Calculate $S_r^t$ using one of scoring functions (\ref{eq:TotalDistance})--(\ref{eq:WeightedNormalizedInverseFrequency})
    \EndFor
  \EndFor
  \State $t^*,r^* \gets (t \in T, r \in R^t) \text{ that maximize } S_r^t \text{ and satisfy } (|R^t| > 1) \text{ or } (\exists t' \in T | W^{t'}_t \neq 1)$ \label{alg:1Phase:FindBestResult}
  \State $R^{t_p^*} \gets \{ r^* \}$ \label{alg:1Phase:ChooseResult}
  \ForAll{$t \in T$} \label{alg:1Phase:StartRemoveTerms}
    \If{$W_t^{t^*} = 0$}
    	  \State $T \gets (T \setminus \{ t \})$
    \EndIf
  \EndFor \label{alg:1Phase:EndRemoveTerms}
  \State Update $P$ to reflect changes in $T$
  \State Recalculate $W^{t_1}_{t_2} \text{ } \forall t_1,t_2 \in T$ as in step 1 \label{alg:1Phase:RealculateWeight}
\EndWhile
\end{algorithmic}
\end{algorithm}

Step \ref*{alg:1Phase:while} repeats while there are still terms that need to be disambiguated: there are still phrases with multiple results, or there are still conflicting (i.e., overlapping) terms.
Step \ref*{alg:1Phase:FindBestResult} is the key step in our greedy algorithm. It finds the term $t^*$ and result $r^*$ that have the highest score $S_{r^*}^{t^*}$. The greedy algorithm selects the highest score, which is considered the likely most coherent match with other terms and results.
(In the case of a tie, the importance that Nominatim assigned to the results is used as the tie-breaker.) Step \ref*{alg:1Phase:FindBestResult} only considers terms that need to be disambiguated: multiple results still exist for these terms, or the terms still overlap with at least one other term. 

After step \ref*{alg:1Phase:FindBestResult} finds the result $r^*$ with the best score, step \ref*{alg:1Phase:ChooseResult} makes $r^*$ the only result that is considered for phrase $t_p^*$. Also, steps \ref*{alg:1Phase:StartRemoveTerms} to \ref*{alg:1Phase:EndRemoveTerms} remove any terms that conflict with $t^*$, settling on the interpretation that contains term $t^*$. This means that after step \ref*{alg:1Phase:EndRemoveTerms} the term $t^*$ has been completely disambiguated, and the algorithm is one step closer to disambiguating all terms in the text.

The other version of the disambiguation algorithm is called the 2-phase disambiguation algorithm, which is described in Algorithm \ref{alg:2Phase}. The difference between the 1-phase and 2-phase versions is that the 2-phase algorithm attempts to first resolve all conflicting terms (i.e., it first settles all interpretation choices), and then disambiguates between location search results in a second phase.

Instead of one while-loop as in Algorithm \ref{alg:1Phase}, there are now two while-loops corresponding to the two phases. The first loop from steps \ref*{alg:2Phase:While1Start} to \ref*{alg:2Phase:While1End} reduces the set $T$ until there are no more conflicting terms. The second loop from steps \ref*{alg:2Phase:While2Start} to \ref*{alg:2Phase:While2End} picks a location search result for each remaining term. 
The 2-phase algorithm cleanly separates the disambiguation of interpretations from the disambiguation of possible location results. The weights are only used to disambiguate interpretations, since all weights are one in the second phase, where location results are disambiguated. In the second phase, all location results for all the terms that remain after the first phase are still considered, and the distance functions that are used to select the location results for each term only depend on actual distances and are not biased by weights or ambiguities coming from unresolved interpretations. Our tests in Section \ref{ch:Results} will investigate whether there are advantages to this approach.

\begin{algorithm}
\caption{2-Phase disambiguation algorithm}
\label{alg:2Phase}
\begin{algorithmic}[1]
\State Calculate $W^{t_1}_{t_2} \text{ } \forall t_1,t_2 \in T$ using Equations (\ref{eq:Weight:Conflict}), (\ref{eq:Weight:SameTerm}), and (\ref{eq:Weight:othergroup})
\While{$\exists t_1,t_2 \in T | (W^{t_1}_{t_2} \neq 1 )$} \label{alg:2Phase:While1Start}
  \ForAll{$t \in T$}
    \ForAll{$r \in R^t$}
      \State Calculate $S_r^t$ using one of scoring functions (\ref{eq:TotalDistance})--(\ref{eq:WeightedNormalizedInverseFrequency})
    \EndFor
  \EndFor
  \State $t^*,r^* \gets (t \in T, r \in R^t) \text{ that maximize } S_r^t \text{ and satisfy } \exists t' \in T | W^{t'}_t \neq 1$
  \ForAll{$t \in T$}
    \If{$W_t^{t^*} = 0$}
    	  \State $T \gets (T \setminus \{ t \})$
    \EndIf
  \EndFor
  \State Update $P$ to reflect changes in $T$
  \State Recalculate $W^{t_1}_{t_2} \text{ } \forall t_1,t_2 \in T$ as in step 1
\EndWhile \label{alg:2Phase:While1End}
\While{$\exists p \in P | (|R^p| > 1)$} \label{alg:2Phase:While2Start}
  \ForAll{$t \in T$}
    \ForAll{$r \in R^t$}
      \State Calculate $S_r^t$ using one of scoring functions (\ref{eq:TotalDistance})--(\ref{eq:WeightedNormalizedInverseFrequency})
    \EndFor
  \EndFor
  \State $t^*,r^* \gets (t \in T, r \in R^t) \text{ that maximize } S_r^t \text{ and satisfy } |R^t| > 1$
  \State $R^{t_p^*} \gets \{ r^* \}$
\EndWhile \label{alg:2Phase:While2End}
\end{algorithmic}
\end{algorithm}

\subsection{Full geolocation algorithm}
\label{sec:AlgSummary}

Now that each step in Algorithm \ref{alg:TechnicalOverview} has been explained in more detail, we can write a more comprehensive summary of the full geolocation algorithm. This is given by Algorithm \ref{alg:Summary}.
Note that the filtering in steps 4--18 to obtain the relevant term set $T$ can be modified and possibly be made more efficient depending on the type of document one wants to geolocate. However, the filtering steps in Algorithm \ref{alg:Summary} are organized in such a way that the geolocation algorithm can be applied to a broad class of texts, including Wikipedia articles, Twitter messages, Kijiji classified ads, etc.

At the end of Algorithm \ref{alg:Summary} we have a collection of terms that are mentioned in the text along with a single location for each. Scores can be recalculated for these results using the same scoring function that was used for the disambiguation. (Note that all weights will be 1 at this point.) They can then be sorted in order of decreasing score to provide a ranking for the top places mentioned in the text. This is done in Section \ref{ch:Results} to compare the best results to the true location for articles.

\begin{algorithm}
\caption{Full geolocation algorithm}
\label{alg:Summary}
\begin{algorithmic}[1]
\State Tag the input text using the POS and NER taggers from \cite{StanfordPOS2000,StanfordPOS2003,StanfordNER}.
\State $S \gets$ ordered list of words in the text
\State $T \gets \emptyset$
\For{every $s \subseteq S$ where $s$ is an ordered sequence of adjacent words}
  \If{$s$ contains at least one word tagged as a NER LOCATION or POS noun}
    \If{every word in $s$ is tagged as noun, adjective, number, or LOCATION}
      \State $t \gets$ a term made from the words in $s$
      \State $T \gets (T \cup {t})$
    \EndIf
  \EndIf
\EndFor
\If{$T$ contains terms that have words tagged with a NER LOCATION}
  \State Remove terms from $T$ that do not contain any words tagged with LOCATION.
\Else
  \If{$T$ contains terms that occured after POS prepositions}
    \State Remove terms from $T$ that did not occur after prepositions.
  \EndIf
\EndIf
\For{every $s \in S$ where $s$ matches a Canadian, American, or Dutch postcode format}
  \If{$s$ is not represented by any term in $T$}
    \State $t \gets$ a term made from the words in $s$
    \State $T \gets (T \cup {t})$
  \EndIf
\EndFor
\State $P \gets $ set of phrases represented in $T$
\For{every $p \in P$}
  \State $R^p \gets$ search results from the knowledge base for query $p$
  \If{$|R^p| = 0$}
    \For{every $t \in T$}
      \If{$t_{phr} = p$}
        \State $T \gets (T \setminus \{ t \} )$
      \EndIf
    \EndFor
  \EndIf
\EndFor
\State Run Algorithm \ref{alg:1Phase} for 1-phase or Algorithm \ref{alg:2Phase} for 2-phase disambiguation
\State \textbf{Return} set of non-conflicting terms $T$ and one result in $R^p$ for every $p \in P$.
\end{algorithmic}
\end{algorithm}

\section{Results for Wikipedia Test Data}
\label{ch:Results}

In this section we describe the performance of our geolocation algorithm with each of the eight scoring functions and both disambiguation algorithms that were discussed in Section \ref{sec:Disambiguation}, for a dataset of Wikipedia articles.

\subsection{Test Data}

We tested our geolocation algorithm with a subset of English geotagged Wikipedia articles, that is, articles about topics with a well-defined geographical location where the authors have provided latitude and longitude coordinates of the primary ``true location'' the article describes. This data required a significant amount of preprocessing, which is described in Appendix \ref{AppendixB}.
Our full dataset contained 920,176 geotagged Wikipedia articles, but the tests in this section used only a sample of 5,976 articles. This subset was used due to time and resource constraints (in particular, the one query per second limit on the free Nominatim servers which was discussed in Section \ref{sec:Nominatim}). The subset was sufficiently large to thoroughly test the capabilities of our geolocation algorithm, as indicated below.

\subsection{Filtering by NER and POS Tags}
\label{sec:Results:Filtering}

Section \ref{sec:LocationExtraction} described how we filter our set of initial terms to come up with the term set $T$ that we disambiguate. See also Algorithm \ref{alg:Summary} (steps 1--24) for a summary of the filtering steps. If the text contains NER LOCATION tags, then we only use terms that have at least one word with this tag (steps 12-13). If there are no LOCATION tags then we only keep terms that occur after POS prepositions (steps 15--17). If there are also no terms that follow prepositions, then we maintain the initial term list based on POS nouns (steps 4--11).

We ran our geolocation algorithm on 5,976 articles. Of these articles, we rejected 500 outright because the different text parsers used by the NER and POS taggers produced inconsistent tokens.\footnote{Note that this does not represent a limitation of our approach. It is rather a minor technical issue that arises because the data was not sufficiently cleaned by the opensource WikiExtractor program we used to pre-process Wikipedia raw data dumps, see Appendix \ref{AppendixB}. This happened, for example, when html tags (like ``$<$br$>$") were not properly removed. In principle, the tokenizers of the NER and POS taggers could be made to match, or, alternatively, efforts could be made to clean the data further and any remaining inconsistencies could be reconciled in a suitable ad-hoc fashion. We chose to simply reject articles that produced these inconsistencies, because we had a very large number of articles that produced consistent NER and POS tokens (many more than we needed for our tests).}
Table \ref{table:filter} shows how often each of the filtering methods were used for the 5,476 articles we considered in our tests. Over 99\% of articles were able to use terms with NER LOCATION tags only, and only rarely did our geolocation algorithm need to fall back on POS prepositions or POS-noun-based terms. This is not surprising for Wikipedia articles that are geotagged by their authors, because they presumably discuss a well-defined location, and are thus likely to contain a good number of terms that the NER may recognize as a LOCATION. As such, these Wikipedia tests will mostly investigate the performance of Algorithm \ref{alg:Summary} for cases where the algorithm relies solely on terms recognized by the NER as LOCATIONs. However, the approach of Algorithm \ref{alg:Summary} is more general and was developed to also be able to handle shorter or unstructured texts with fewer clear NER LOCATION terms, by relying on POS prepositions and/or nouns. These capabilities of our approach will be tested in Section \ref{ch:Results-Twitter}.

\begin{table}
\begin{tabular}{|c|c|c|}
\hline
\textbf{Terms Used} & \textbf{\# of Articles} & \textbf{Fraction of Articles} \\
\hline
NER LOCATION tags only & 5446 & 99.45\% \\
\hline
Terms after POS prepositions & 28 & 0.51\% \\ 
\hline
Initial term list based on POS nouns & 2 & 0.04\% \\
\hline
\end{tabular}
\caption{Terms used for disambiguating Wikipedia test articles (see Algorithm \ref{alg:Summary}, steps 1--24).}
\label{table:filter}
\end{table}



\subsection{Comparison of Disambiguation Algorithms}
\label{sec:Results:AlgComparison}

\begin{figure}
\includegraphics[width=\textwidth]{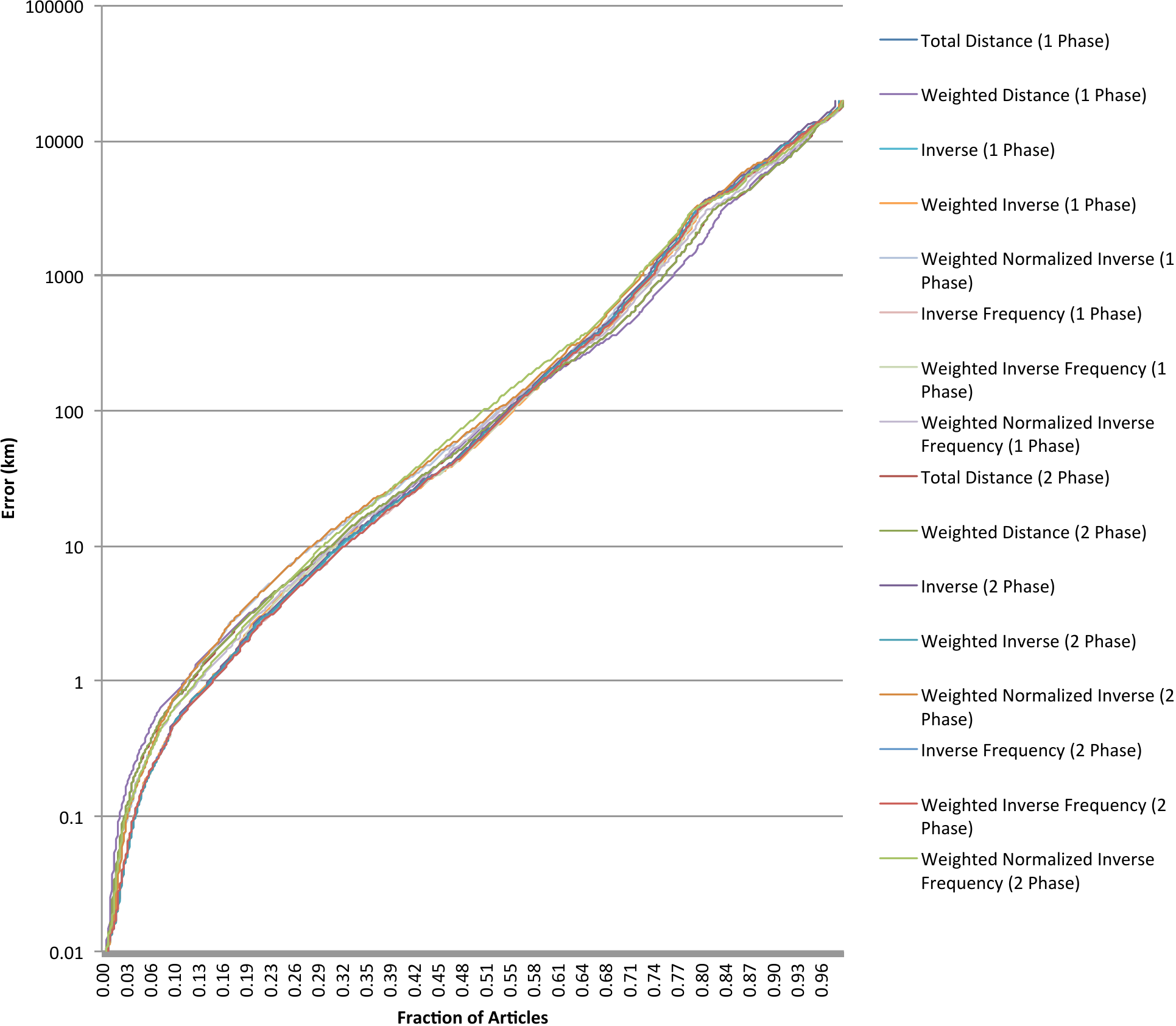}
\caption{Wikipedia tests. Error for disambiguation algorithms.}
\label{fig:DisamgibuatorComparison}
\end{figure}

\begin{figure}
\centering
\includegraphics[scale=0.65]{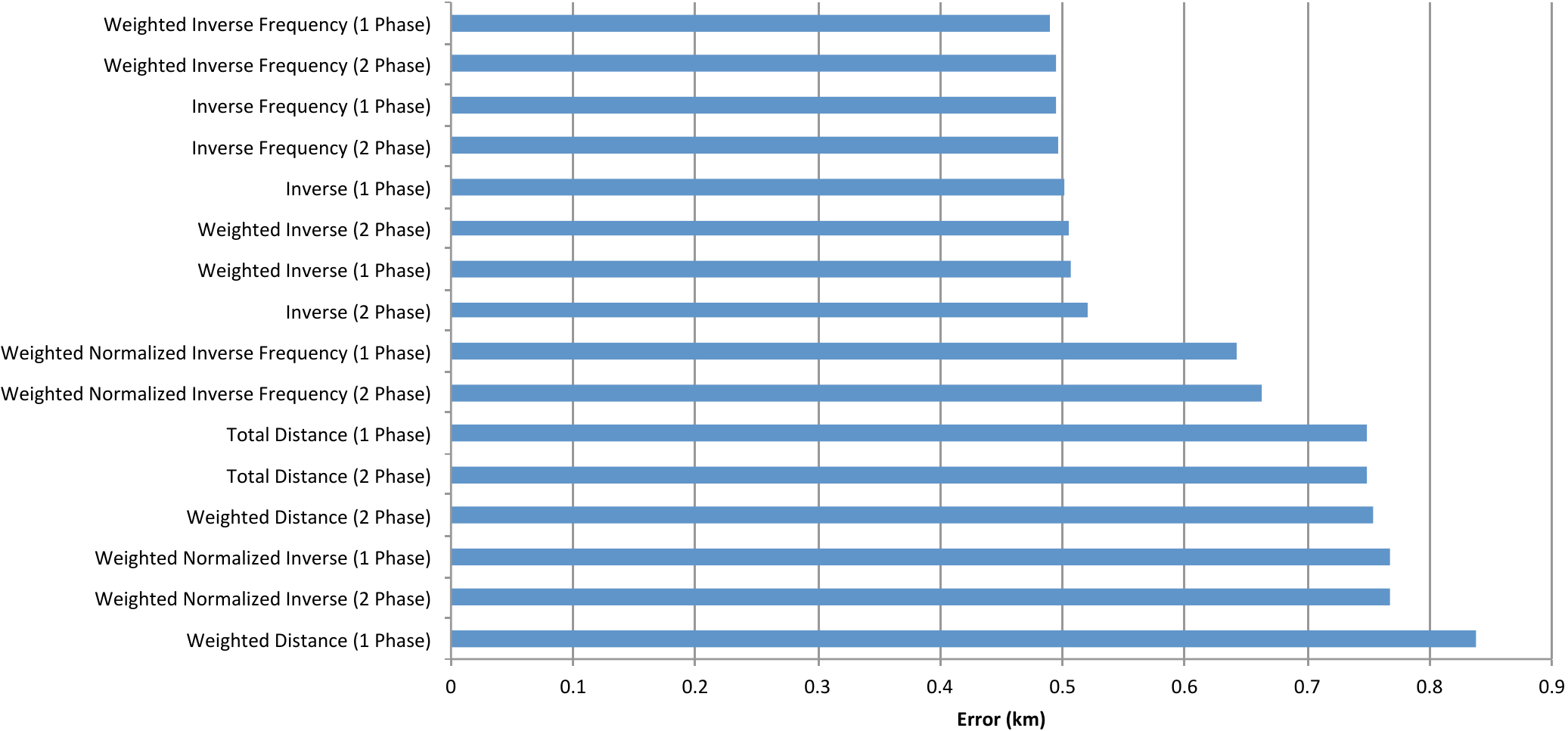}
\caption{Wikipedia tests. Error for disambiguation algorithms at the 10th percentile.}
\label{fig:Error10}
\end{figure}

\begin{figure}
\centering
\includegraphics[scale=0.65]{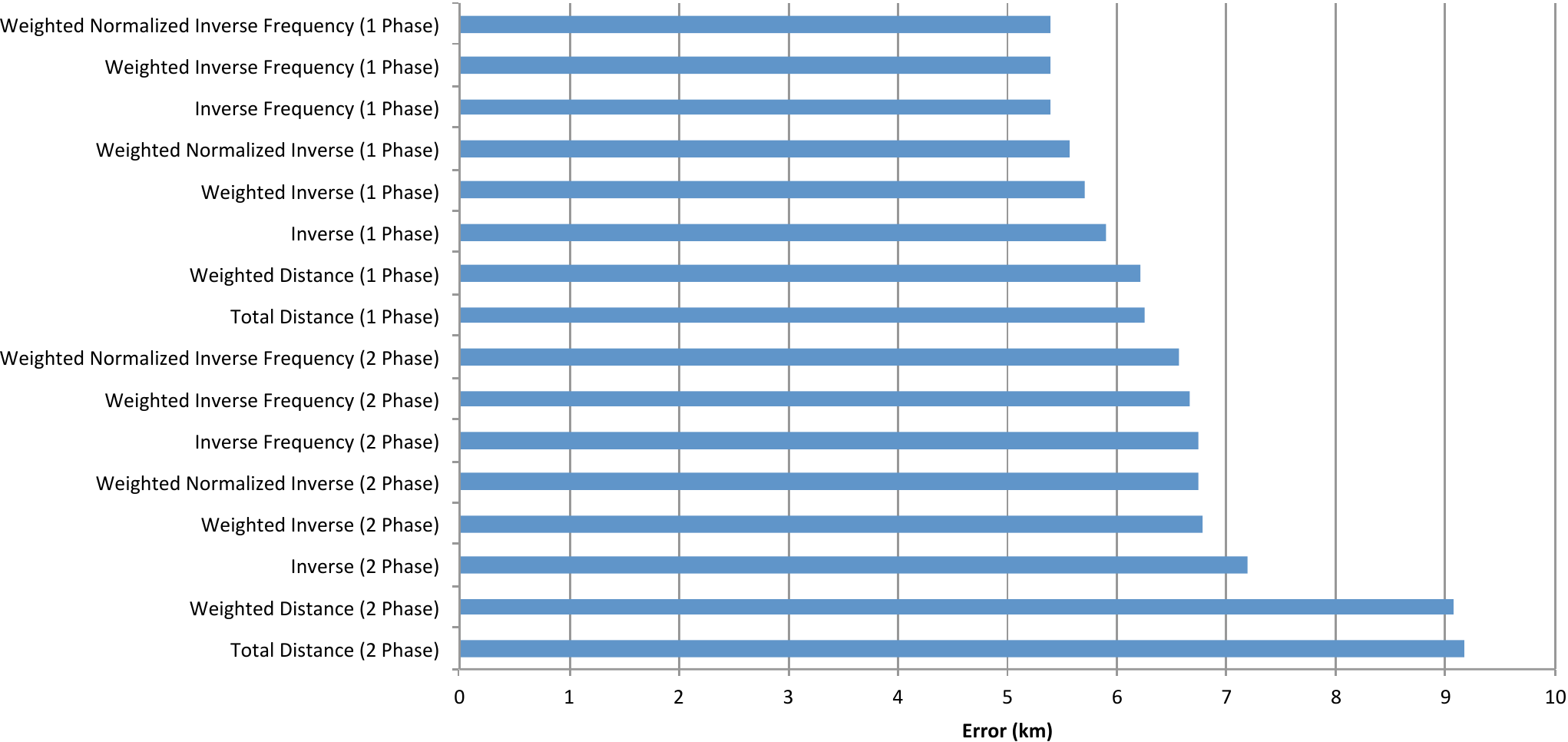}
\caption{Wikipedia tests. Error for disambiguation algorithms at the 25th percentile.}
\label{fig:Error25}
\end{figure}

\begin{figure}
\centering
\includegraphics[scale=0.65]{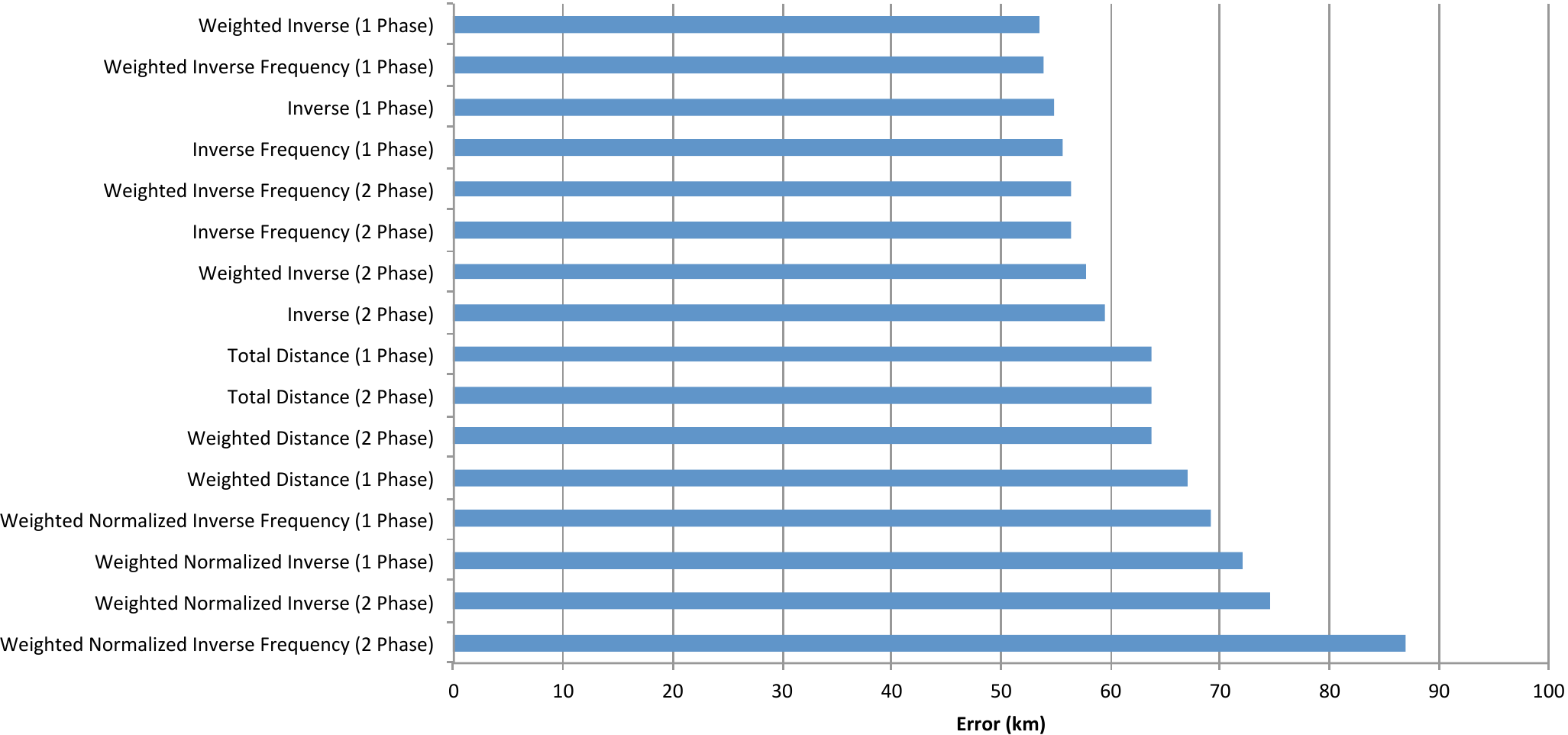}
\caption{Wikipedia tests. Error for disambiguation algorithms at the 50th percentile.}
\label{fig:Error50}
\end{figure}

Our next step was to determine which versions of the geolocation algorithm are the most accurate. There are eight scoring functions and two disambiguation algorithms, giving a total of 16 versions to compare.
Each geotagged Wikipedia article in our data set has one set of coordinates which is considered to be the true location for the article. Our geolocation algorithm provided as final output a list of geolocated terms with one most likely Nominatim result each. This means we needed to compare our list of multiple locations to the one true location for the article. We first measured error by calculating the distance between the true location and the result with the highest distance-based score, where scores were calculated using the same scoring function that the disambiguation algorithm used. In Section \ref{sec:Results:Top5} we will use a different error measurement that takes multiple ranked outputs into account.
Articles were sorted in order of increasing error for each version of the geolocation algorithm, and Figure \ref{fig:DisamgibuatorComparison} plots the cumulative distribution of the error (in km) obtained for the articles, with the $x$-axis indicating the fraction of articles that have an error below the error value indicated on the logarithmic $y$-axis.
For example, this graph tells us that all versions of our geolocation algorithm had an error of less than one kilometre for at least 11\% of articles.
Note that the horizontal axis of Figure \ref{fig:DisamgibuatorComparison} does not reach 100\%. This is because disambiguation algorithms were terminated when their execution time exceeded 100 seconds. Out of 87,616 disambiguation attempts (16 algorithm versions times 5476 articles), this cutoff was used 496 times (0.57\%).
We expected this graph to show one or two algorithms that were clearly better than the others, but this is not the case. All algorithms showed similar results, and there appeared to be no clear winner. So instead we sliced the graph at the 10th, 25th and 50th percentiles, as shown in Figures \ref{fig:Error10}, \ref{fig:Error25}, and \ref{fig:Error50}, respectively.
This confirmed that there was no algorithm which was a clear winner. However, the 1-phase algorithm with the Weighted Inverse Frequency scoring function (Equation (\ref{eq:WeightedInverseFrequency})) was one of the two best algorithms in each of the three figures. Thus we chose it as our winning algorithm, and we will study it further in Section \ref{sec:Results:Winner}.

\subsection{Further Analysis with the Winning Algorithm}
\label{sec:Results:Winner}

We now consider test results for the Weighted Inverse Frequency scoring function (Equation (\ref{eq:WeightedInverseFrequency})) with the 1 phase disambiguation algorithm (Algorithm \ref{alg:1Phase})
in more detail.

\paragraph{Results by Article Type}

The author-provided location tags for Wikipedia articles each have a type associated with them. We investigated whether some of these types are more difficult for the algorithm to geolocate than others.
Nineteen different types of tags were observed in the test set, in addition to a NULL (or missing) type. Ten of these types, including the NULL type, occurred at least 50 times in the 5476 articles that were tested. Table \ref{table:Results:ArticleType} shows the median error (50th percentile) for each of these types.
The table shows that airports and railway stations appear to be the easiest to geolocate, as they have the lowest median error. These types of articles are likely to name both the location and the nearby cities they serve, so our algorithm can expect to find a small cluster of results with good scores resulting in easy disambiguation.

In contrast, rivers and second-level administrative regions (districts, counties, etc.) appear to be the most difficult to locate. This makes sense, as these types of locations can be spread across a large area but are only represented by a single point in the Wikipedia ``true location''. We use only the single points that are associated with locations in the OpenStreetMap knowledge base, and we compute distances between these single points. There is no pre-specified way for a Wikipedia author to assign a single location to a large geographic entity (like a state or province, or, even more so, an essentially one-dimensional entity like a river), and it is not to be expected that Wikipedia authors assign locations in a way that is consistent with OpenStreetMap locations, for these types of large geographic entities. It is therefore to be expected that large deviations occur between the Wikipedia ``true location'' and what our geolocation algorithm can obtain based on OpenStreetMap (and deviations of this type can, in fact, hardly be called ``errors'').
For example, if an article is about a river and our algorithm returns a location beside the river, this could show very poor accuracy if the returned point is far from the centre point that is assigned to the river by OpenStreetMap.
One possible way to improve this in future work is to consider a radius for geographic objects in addition to a centre point. If Wikipedia articles and OpenStreetMap entities were tagged with a (centre point, radius) pair, we could use shortest distances between discs in our distance-based scoring functions and in our error measures. 
For essentially one-dimensional entities like rivers, the geometric moments could be considered to augment the
location description. These improvements are a daunting task since the required data is not readily available, but it could certainly be expected to dramatically reduce the perceived error of geolocation algorithms like the one proposed in this paper, for geographic entities that are large or essentially one-dimensional. 

\begin{table}
\begin{tabular}{|c|c|c|}
\hline
\textbf{Article Type} & \textbf{\# of Articles} & \textbf{Median Error (km)} \\
\hline
airport & 60 & 6 \\
\hline
railwaystation & 178 & 6 \\
\hline
waterbody & 76 & 42 \\
\hline
city & 2414 & 44 \\
\hline
mountain & 124 & 54 \\
\hline
landmark & 1109 & 65 \\
\hline
NULL & 970 & 94 \\
\hline
edu & 187 & 110 \\
\hline
river & 86 & 187 \\
\hline
adm2nd & 105 & 295 \\
\hline
\end{tabular}
\caption{Wikipedia tests. Accuracy comparison for different article types.}
\label{table:Results:ArticleType}
\end{table}

\paragraph{Confidence Estimation}

\begin{figure}
\centering
\includegraphics[scale=.8]{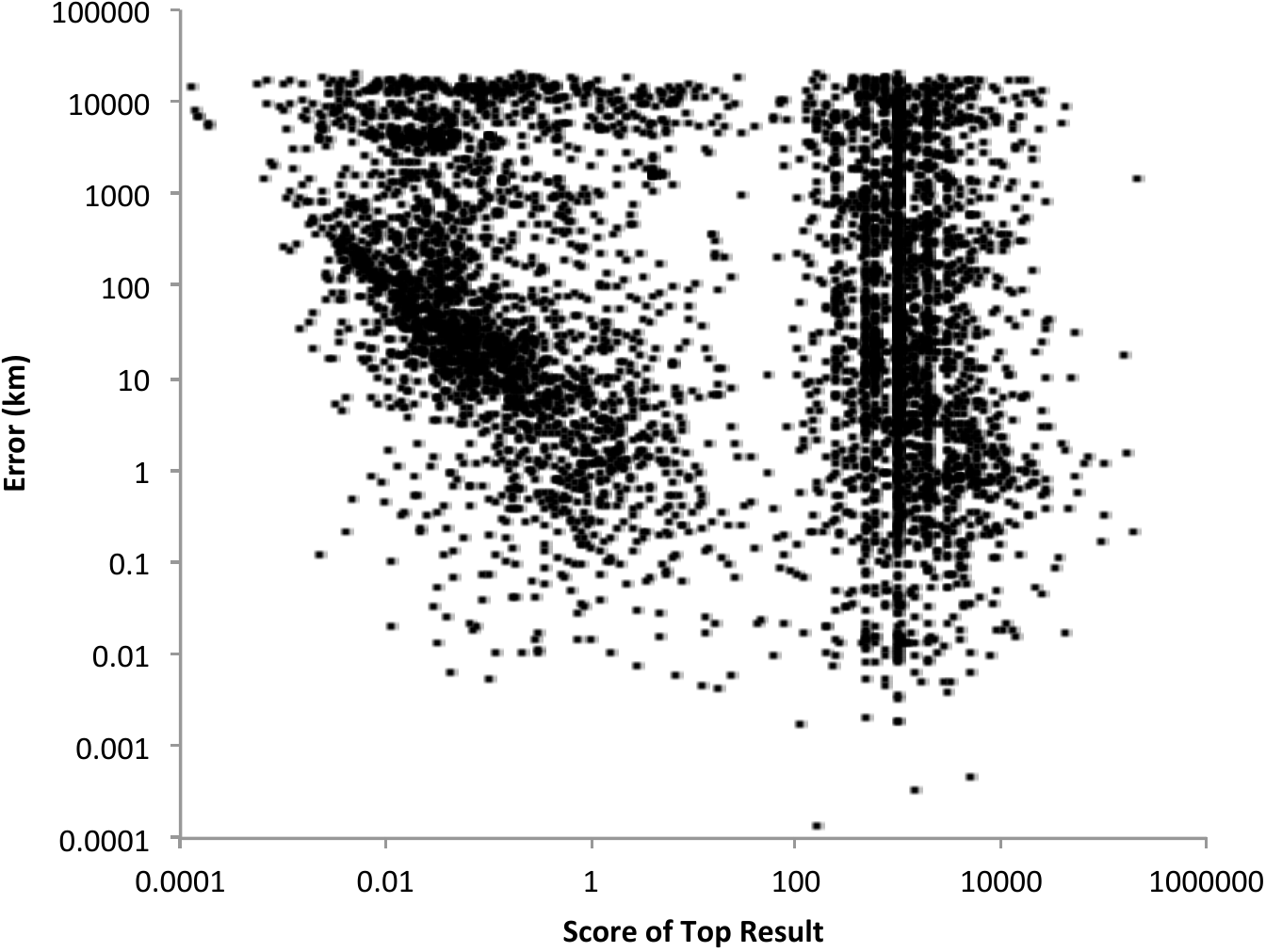}
\caption{Wikipedia tests. Error versus scores for weighted inverse frequency (1 phase).}
\label{fig:WIFscatter}
\end{figure}

A valuable addition to our geolocation algorithm would be an estimator for the accuracy of the algorithm when the true error is unknown. This would allow us to assign a confidence value to each location tag.
Figure \ref{fig:WIFscatter} shows the error in the location tagging as a function of the score of the top result. It could be expected that there would be a clear downward trend, where larger scores correlate to smaller errors. This would give us a simple relationship where the score of the top result would also be our confidence measure. However, our results did not indicate such a relationship and, as shown in Figure \ref{fig:WIFscatter}, there were three main clusters instead.
The bottom left cluster does show some of the desired trend, where larger scores indicate lower error. However, this is not simple to interpret, as there is another cluster directly above with a horizontal shape.
The third cluster is on the right, and spans the entire $y$-axis range of the graph. This cluster shows some vertical stratification starting around 1000 km. This is likely an artefact of the maximum we imposed on the denominator of Equation (\ref{eq:WeightedInverseFrequency}), where any distances less than 1 metre were increased to 1 metre as an attempt to avoid division by zero. Every time this restriction is imposed the score increases by 1000. This is likely the major cause of the stratification that is observed around the integer multiples of 1000 in Figure \ref{fig:WIFscatter}.

When the top result produced by our geolocation algorithm does not correspond to the ``true location'' of the Wikipedia article, the error can easily be large to very large. In this case, the score of the top result for the weighted inverse frequency distance function can be very large if the top result is very close to other results (which may produce the multiple-1000 stratification in the right cluster of Figure \ref{fig:WIFscatter}), or it can be moderately large if the top result is moderately close to other results (the top cluster in Figure \ref{fig:WIFscatter}). Large geographical entities could also easily lead to errors that appear much larger than they are in reality. Even though Wikipedia articles may have a single location tag, they can be expected to often contain (clusters of) terms referring to locations that are not close to the single location tag, just because Wikipedia articles typically contain a large number of words and describe their topics in a broad context. This may also easily lead to large errors in geolocation based on NER and POS combined with our distance-based scoring functions, for some geotagged Wikipedia articles.

The results indicated in Figure \ref{fig:WIFscatter} do not give us a simple answer in our search for a confidence measure. A few other confidence measures were tried with similar results, including the ratio between the top two scores, as well as the score multiplied or divided by the number of terms used to calculate the score. None of these showed a clearer relationship, so we do not yet have a confidence measure for our geolocation algorithm, and it may be difficult to obtain one that is highly reliable, given the difficulty of the geolocation problem (for a fraction of Wikipedia articles our approach works very well, but there will likely always be a fraction of difficult cases for which it does not work well). 
For example, we replotted the error in location tagging as a function of the score of the top result using the Weighted Distance (1-phase) version of the algorithm (plot not shown), and this did not show the stratification that was apparent in Figure \ref{fig:WIFscatter}. However, there was still no clear relationship between score and error. Future work includes investigating all 16 versions, although given the similarities between them it is unlikely that this will result in a reliable confidence measure.

\subsection{Analysis of Top 5 Results}
\label{sec:Results:Top5}

In previous sections we used the distance between the true location and the result with the greatest score as a measurement of error. This may not be the best approach, as our algorithm provides more information in a list of multiple locations mentioned in the text. So far we used only one of these locations to calculate the error.
For example, if a text describes a river and mentions the cities that lie along it, then the geolocation algorithm would produce locations for the cities as well as a location for the river. The ``true location'' for the article would be a single point somewhere along the river. If one of the cities on the river has the highest score in our result set, and this city happens to be far from the point assigned to the river by the author of the article, then this may give a very large error value even though our output may be considered accurate.
It may make more sense for a case like this to consider a group of locations that are returned by the geolocation algorithm. We therefore redefined our error measure to be the shortest distance between the ``true location'' and any of the five locations produced by our geolocation algorithm with the greatest scores, rather than just the top location with the greatest score.

\begin{figure}
\includegraphics[width=\textwidth]{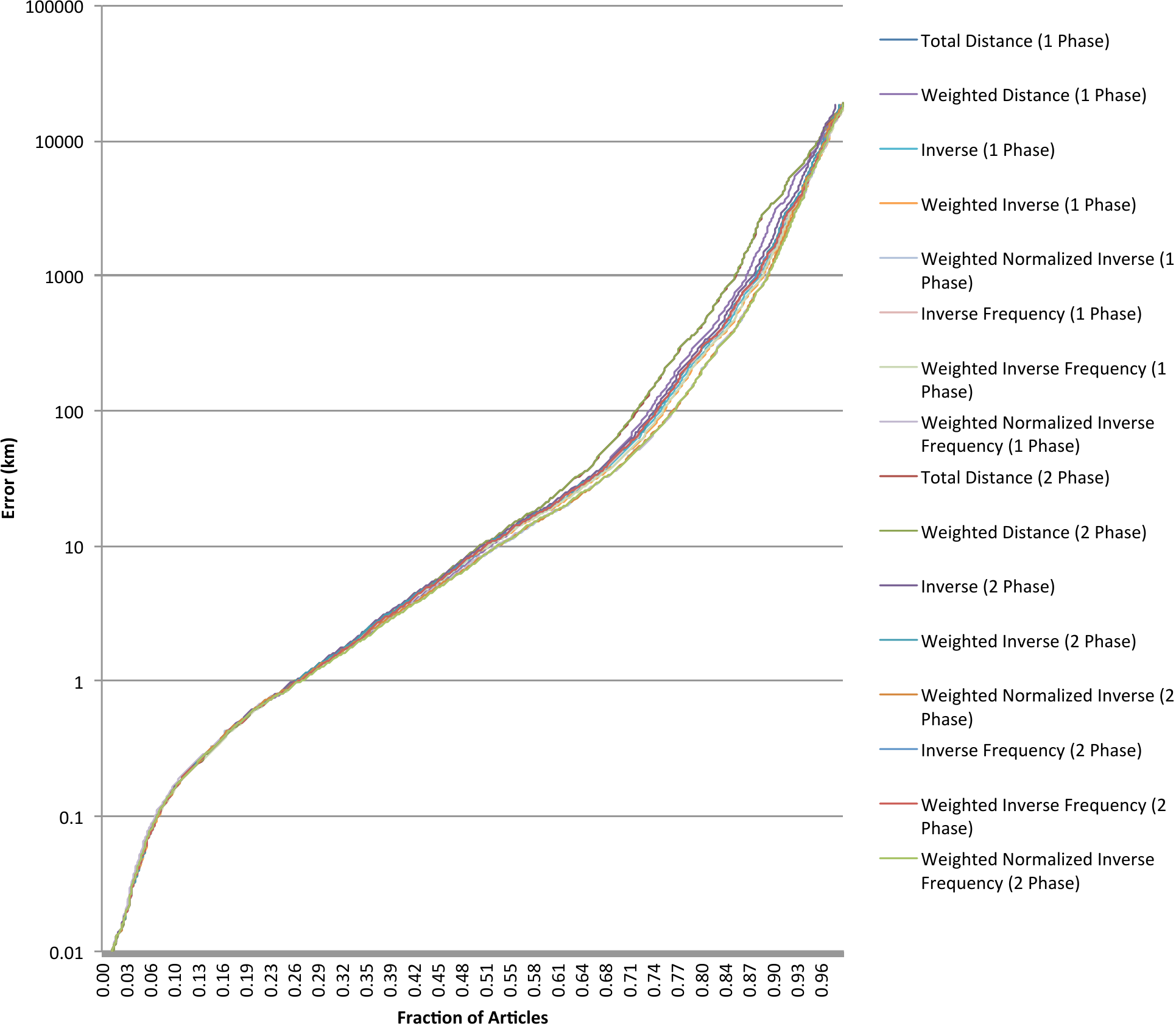}
\caption{Wikipedia tests. Top 5 error for disambiguation algorithms.}
\label{fig:DisamgibuatorComparisonTop5}
\end{figure}

Figure \ref{fig:DisamgibuatorComparisonTop5} is a modified version of Figure \ref{fig:DisamgibuatorComparison} which uses this Top 5 error measurement. Here we see the curves converge closer to each other than in Figure \ref{fig:DisamgibuatorComparison}, meaning the 16 algorithms are more similar in this case. We also see a significant improvement in error. For example, the Weighted Inverse Frequency (1 phase) algorithm had a 10th percentile error of 490 metres with the Top 1 error, while for the Top 5 this error is 163 metres. Similarly, the error at the 25th percentile improved from 4.70 km to 0.87 km. Finally, the median error improved from 54 km to 8.5 km.
This new error measurement makes our median error similar to those observed by Roller et al. \cite{Roller} using a classification approach.
However, this error measure has access to the true answer can and pick the top 5 result that is closest to it. This does tell us that we might significantly improve our accuracy by using a more-informed scoring function to rank the final results. This is left for future work.
Recall also that the maximum accuracy of classification-type approaches is effectively limited by the class size, and the accuracies we achieve for large numbers of articles (e.g., 0.87 km error at the 25th percentile) cannot be attained by classification approaches.

\section{Results for Twitter Test Data}
\label{ch:Results-Twitter}

We now turn to tests with Twitter data, to further evaluate the performance of our geolocation algorithm.
We consider geotagged tweets, i.e., tweets that contain the geographic location from which they were sent in their metadata. We will attempt to geolocate these tweets (or concatenations of tweets sent out by the same user) solely based on the text in the tweets, and will assess which fraction of the tweets (or concatenations) are geolocated by our algorithm to a location close to the location from which the tweets were sent (which we consider as the ``true location'' for assessing our geolocation algorithm for tweets). Geolocating tweets is inherently challenging: tweets are short and unstructured, and are less likely to contain NER LOCATIONs. This means that our geolocation algorithm will more often need to rely on POS prepositions and nouns (see Algorithm \ref{alg:Summary}). Moreover, unlike the Wikipedia articles about well-defined geographic entities that were specifically geotagged by their authors, the contents of most geotagged tweets may actually not at all be about the location from which the tweet was sent. We thus expect our algorithm to geolocate few tweets to a location close to the location from which the tweets were sent. In fact, in experiments reported below we estimate an upper bound on the fraction of geotagged tweets that can in principle be geolocated to the location they were sent from based on information in the tweet text. Regardless of these limitations, the tests reported below are an interesting way to assess the possible applicability of our approach for short and unstructured text messages, and as such complement the Wikipedia results discussed in Section \ref{ch:Results}.

\subsection{Test Data}
We have compiled a Twitter dataset containing nearly all geotagged tweets from Canada, the US, the Netherlands, and Australia for the period of February, March, and April 2015. The dataset contains Tweets from 3,764,507 unique users. The dataset was collected using the Twitter API via the Spark framework for data analytics, as explained in Appendix \ref{AppendixC}.

\begin{figure}
\centering
\includegraphics[width=0.85\textwidth]{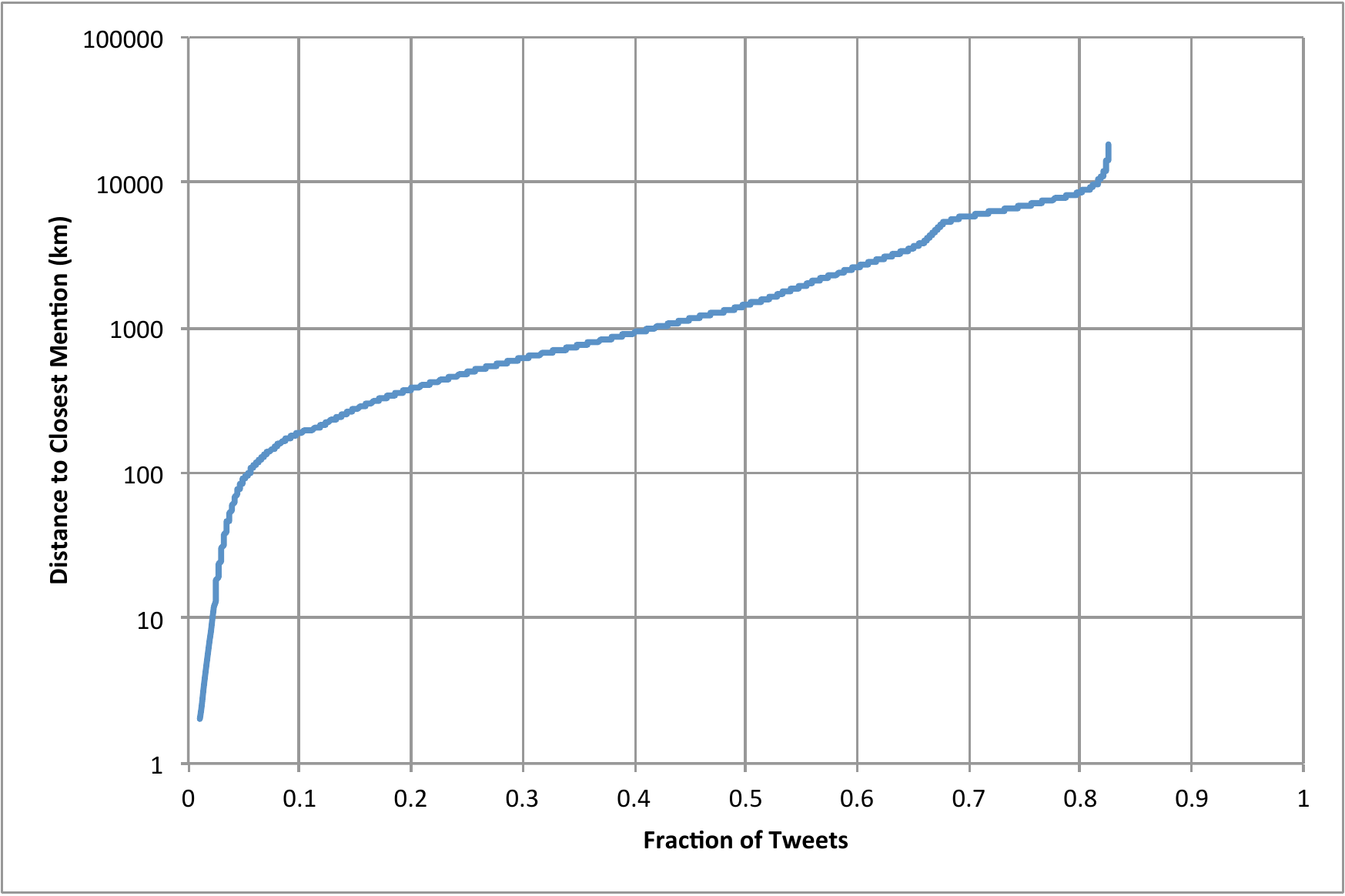}
\caption{Twitter tests. Cumulative plot of the distance between the coordinates of a sample of geotagged tweets, and the closest mention of a location in the tweet text (that is present in the GeoNames knowledge base), as a function of the fraction of tweets.}
\label{fig:GeoNames}
\end{figure}

It is interesting to first consider what fraction of geotagged tweets contain words or sequences of words that can be interpreted (using a knowledge base) as locations within a reasonable distance of the location the tweet was sent from. In particular, we considered a 10\% sample of the February 2015 tweets we collected, consisting of 3,466,922 geotagged tweets from our larger dataset. For each tweet we considered all word sequences with up to 5 words (no POS or NER was used; we just considered every possible sequence of 5 or less consecutive words). We then did a case-insensitive search through all locations in the GeoNames knowledge base (\url{http://www.geonames.org}).
This search included ``alternative names'' for locations with more than one name. We measured the distance between each match and the coordinates of the tweet, and kept the shortest distance for each tweet. Figure \ref{fig:GeoNames}
shows a cumulative plot of the distance between the tweet coordinates and the closest mention in the tweet text, as a function of the fraction of tweets. We see immediately that few tweets contain location references in the tweet text that are close to the tweet coordinates. In particular, we find that
\begin{itemize}
\item 82.5\% of tweets contain a word sequence that matches a location in GeoNames
\item 2.2\% of tweets mention a location within 10 km
\item 5.4\% of tweets mention a location within 100 km
\item 8.3\% of tweets mention a location within 161 km (100 miles)
\end{itemize}
This shows that we cannot expect high accuracy when attempting to geolocate tweets and comparing to the location the tweet was sent from. Nevertheless, it is interesting to see whether our approach manages to correctly geolocate tweets that do contain relevant location information in the text.
(Note that we used GeoNames instead of OpenStreetMap for this particular test, since the GeoNames data was in a more practical format for our current purposes. The general points we made here do not depend on whether one uses GeoNames or OpenStreetMap.)

\subsection{Results}
Since we cannot expect high recall accuracy for single tweets, we first consider concatenations of tweets sent by the same user, which may have more useful NER LOCATIONs or POS nouns/prepositions to base geolocation on.

In our first set of tests, we combined, for every user, the text of all the user's tweets into one text.
For each user, we set the user's latitude and longitude to be the mean latitude and longitude of all the user's tweets in the dataset. We considered this for a sample of 882 users, and find the following results:
\begin{itemize}
\item 10th percentile: 603 km
\item 25th percentile: 2,222--2,247 km (note: we report the range for the 16 different methods)
\item 50th percentile: 6,646--6,665 km
\end{itemize}
Note that, as expected, this compares poorly with the Wikipedia results from Figures \ref{fig:Error10}--\ref{fig:Error50}, where the best algorithm produced an error of about 0.5 km at the 10th percentile, 5.5 km at the 25th percentile, and 53 km at the 50th percentile. It is, however, roughly in line with the GeoNames analysis above, which showed that only 8.3\% of tweets mention a location within 161 km.

It is interesting to consider whether our geolocation algorithm found NER LOCATIONs, or had to rely on POS prepostions or nouns. Table \ref{table:filter-Twitter} shows that NER LOCATIONs were only encountered in 1.47\% of the (combined) texts, in stark opposition to the 99.45\% for geotagged Wikipedia articles, see Table \ref{table:filter}.
Instead, our geolocation algorithm had to rely on POS nouns for 48.07\% of the texts, and on POS prepositions for 33.56\% of the texts. Also, for 9.30\% of the (combined) Twitter texts, terms were considered, but there were no Nominatim results for any of the terms in the term list, and for 7.60\% of the texts no terms were found by Algorithm \ref{alg:Summary}. This shows that geotagged Twitter texts (even when combined) are much harder to geolocate than Wikipedia articles that describe geographic entities.

\begin{table}
\begin{tabular}{|c|c|c|}
\hline
\textbf{Terms Used} & \textbf{\# of Articles} & \textbf{Percent of Articles} \\
\hline
LOCATION tags only & 13 & 1.47\% \\
\hline
Terms after prepositions & 296 & 33.56\% \\ 
\hline
Initial term list based on nouns & 424 & 48.07\% \\
\hline
\hline
Terms, no Nominatim results & 82 & 9.30\% \\
\hline
No terms & 67 & 7.60\% \\
\hline
\end{tabular}
\caption{Twitter tests. Terms used for disambiguating Twitter texts (combined per user) (see Algorithm \ref{alg:Summary}, steps 1--24).}
\label{table:filter-Twitter}
\end{table}


For completeness, we briefly report on test results for single tweets. We considered 18,720 tweets, and find a 10th percentile error in the range 1,042--1,043 km (range for the 16 different methods), a 25th percentile in the range 3,406--3,414 km, and a 50th percentile error in the range 7,882--7,892 km. The results for the case of tweets combined by user that were reported above are somewhat better, as expected.


\section{Discussion}
\label{ch:Conclusion}

This paper has presented a geolocation algorithm for finding precise locations that are named in text. It uses part-of-speech tagging and named entity recognition to find potential location references. It then uses the open-source tool Nominatim to search the OpenStreetMap database and obtain a list of possible locations for each unique phrase. Finally, conflicting terms are resolved and one location is chosen for each phrase. The algorithm returns a list of non-conflicting terms and the location that was chosen for each of them.

Sixteen versions of the algorithm were compared, with similar results. The Weighted Inverse Frequency scoring function (Equation (\ref{eq:WeightedInverseFrequency})) with the 1-phase disambiguation algorithm (Algorithm \ref{alg:1Phase}) performed consistently near the top, and was subsequently used to report detailed accuracy results.

When the location with the highest score was chosen to represent the overall location of the article, our median error was 54 km. This is worse than the 11.0 km reported by Roller et al. \cite{Roller} for a classification approach, but many of our articles are geolocated with much higher precision than what can be achieved by classification approaches (for example, 25\% of articles were geolocated with an error below 5.5 km). Also, when we consider the closest of the five highest-scoring results our median error reduces to 8.5 km. This implies that our algorithm, which is designed to return multiple locations, includes some very accurate results in most cases. Note that future work on the final sorting of results may improve our 54 km median error.


There are many ways in which our geolocation algorithm can be improved in future work. Although the algorithm was developed heuristically, it is a good starting point for a new class of algorithms for high-precision location tagging. 


For example, Habib \cite{Mena} demonstrated that there is a feedback loop between the extraction step and the disambiguation step in named entity recognition. Results from the disambiguation step can be used to improve results from the extraction step, which can then improve results for the disambiguation step. This concept was partially used in this paper. In particular, the extraction step gathered all possible interpretations, then the disambiguation step refined the list of extracted terms by resolving those that conflicted.
Our geolocation algorithm could benefit from more use of this feedback loop. The disambiguation algorithm could be modified to discard terms which have extremely low scores, not just those that conflict with better terms, as these terms may not refer to real locations and should not have been included in the extraction step. This may improve accuracy when disambiguating the other terms in the text.

If our geolocation algorithm is to be applied to a specific type of text, then the algorithm could be improved by using part-of-speech taggers and named entity recognizers that are specifically trained for the target style of text. For example,  the Stanford POS tagger we used could be replaced by the state-of-the-art Twitter POS tagger developed by Derczynski et al. \cite{TwitterPOS} when geolocating tweets.



It may be interesting to consider applying elements of our approach, for example, our distance-based scoring, to more general named entity disambiguation frameworks such as AIDA \cite{hoffart2011robust,yosef2011aida}. This may benefit the accuracy of these frameworks when the entities to be disambiguated include locations.

\appendix

\section{External Software}
\label{AppendixA}

The following software was used in the implementation of our project.
\begin{itemize}
\item Apache Spark version 1.4.0. Based on the work of \cite{Spark}. Available at \url{http://spark.apache.org/}.

\item Nominatim. It was queried using the API described at \url{http://wiki.openstreetmap.org/wiki/Nominatim}. It can also be used through a user interface at \url{https://nominatim.openstreetmap.org/}.

\item PHP-Stanford-NLP. October 18, 2014 version. Written by Anthony Gentile. Available at \url{https://github.com/agentile/PHP-Stanford-NLP}. Used as a PHP-wrapper to access the Stanford POS tagger and NER.

\item Stanford Log-linear Part-Of-Speech Tagger version 3.5.2 (April 20, 2015). Based on the work of \cite{StanfordPOS2000} and \cite{StanfordPOS2003}. Available at \url{http://nlp.stanford.edu/software/tagger.shtml}. There are multiple pre-trained models available with this software, and our project used the model titled \emph{english-left3words-distsim}.

\item Stanford Named Entity Recognizer version 3.5.2 (April 20, 2015). Based on the work of \cite{StanfordNER}. Available at \url{http://nlp.stanford.edu/software/CRF-NER.shtml}. There are multiple pre-trained models available with this software, and our project used the model titled \emph{english.all.3class.distsim.crf}.

\item WikiExtractor. June 14, 2015 version. Written by Giuseppe Attardi. Available at \url{https://github.com/attardi/wikiextractor}.
\end{itemize}

\section{Wikipedia Data}
\label{AppendixB}

At the time of writing, data dumps of the English Wikipedia could be obtained from \url{https://dumps.wikimedia.org/enwiki/}. This paper used articles from a dump of the English Wikipedia on June 2, 2015 (file name enwiki-20150602-pages-articles.xml.bz2). Location tags were also obtained from the June 2, 2015 dump, and are available as a separate download on the same site (file name enwiki-20150602-geo{\textunderscore}tags.sql.gz). Our full dataset contained 920,176 geotagged Wikipedia articles, and our tests used a sample of 5,976 articles.

The raw dump file was processed using WikiExtractor (\url{https://github.com/attardi/wikiextractor}), which takes the XML file and produces the raw text for each article, removing special formatting and annotations such as images, tables, or references. However, the output of WikiExtractor does not contain location tags, so we had to join this data with the set of location tags ourselves.
The geo{\textunderscore}tags file contains all location tags in the English Wikipedia. There are two types of tags: primary and non-primary. Primary tags apply to an entire article. Non-primary tags apply to a particular mention in the text of an article. Our project only used primary tags.\footnote{
It is worth noting that our set of articles is different from the one created by Wing and Baldridge \cite{Wing}, which was also used by Roller et al. \cite{Roller}. They used a dump of the English Wikipedia from September 4, 2010 while we used a dump from June 2, 2015. Their dataset contained 10,355,226 articles (including non-geotagged ones), while ours contained 4,855,711. It should be expected that the more recent data would contain more articles, but the opposite is observed. However, Wing and Baldridge had to explicitly remove redirect articles, giving a total of 3,431,722 content-bearing articles. For this paper we did not explicitly remove such articles, and none have been observed in our dataset. Therefore it is likely that these were automatically excluded in our download or by the processing software we used. Wing and Baldridge used their own processing software to extract article coordinates and obtained 488,269 geotagged articles. We used the set of coordinates that are directly available from Wikipedia.
}

These two data sets were joined using Apache Spark, which is a scalable cluster computing system \cite{Spark}. Spark now has a highly active community of both developers and users. Spark's APIs made it easy to read in both data sets, match location tags to their articles, and output in a clear format using less than 80 lines of Scala code. The output of this Spark program was directly usable by our geolocation algorithm implementation.

\section{Twitter Data}
\label{AppendixC}

Twitter provides an API to allow applications to stream a fraction of the publicly available tweets (\url{https://dev.twitter.com/streaming}). Each application can receive up to approximately 1\% of public tweets, though the sample of tweets that are received can be adjusted with the API. Apache Spark's streaming library (\url{http://spark.apache.org/streaming/}) includes a wrapper for Twitter's API, which allows these tweets to be processed using Spark. A Spark Streaming application was written to receive a stream of tweets and save them to Parquet files for further processing. This program was run for several months to obtain the dataset that was used in Section \ref{ch:Results-Twitter}.

Note that Twitter's API allows applications to specifically request geotagged tweets (i.e., tweets that contain the location from which the tweet was sent as part of the tweet's metadata), but Spark's wrapper for Twitter does not implement this feature. The Spark source code was modified for this paper to allow requests for geotagged tweets within a specified range of coordinates (we specify Canada, the US, the Netherlands, and Australia). Without this, the 1\% sample of tweets would be a random sample of all public tweets on Twitter, and would not specifically be composed of geotagged tweets. By requesting geotagged tweets from these specific regions, the requested sample was less than 1\% of the total amount of public tweets. This meant that the Spark Streaming application collected nearly all public geotagged tweets in those areas.


\section{Code and Data}
\label{AppendixD}

The code for our geolocation algorithm can be obtained at
\url{https://github.com/spotzi/Geotagger}.

The Spark programs we used to pre-process the Wikipedia data and to collect and pre-process the Twitter data can be obtained at \url{https://github.com/sjbrunst/GeoTextTagger-data}. 
The Wikipedia and Twitter data we used for our tests can also be obtained at
\url{https://github.com/sjbrunst/GeoTextTagger-data}.





\bibliographystyle{plos2015}
\phantomsection  

\bibliography{cmreportreferences}

\nocite{*}
\end{document}